\documentclass{bmcart}

\usepackage{hyperref} 
\usepackage{graphicx} 
\usepackage{comment} 
\usepackage{listings} 
\usepackage{xcolor} 
\usepackage{color, colortbl} 
\usepackage{amsfonts}
\usepackage{relsize}
\usepackage{soul}

\definecolor{babyblueeyes}{rgb}{0.63, 0.79, 0.95}
\usepackage[caption=false]{subfig} 
\usepackage{authblk}

\usepackage{makecell}

\startlocaldefs
\endlocaldefs

\begin{document}

\begin{frontmatter}

\begin{fmbox}
\dochead{Regular Article}


\title{Generating Mobility Networks with Generative Adversarial Networks}


\author[
   addressref={aff1},                   
   email={giovanni.mauro@phd.unipi.it},   
   addressref ={aff1, aff2, aff3}
]{\inits{GM}\fnm{Giovanni} \snm{Mauro}}
\author[
   addressref={aff4, aff6},
   email={mluca@fbk.eu}
]{\inits{ML}\fnm{Massimiliano} \snm{Luca}}
\author[
   addressref={aff5, aff6},
   email={alonga@fbk.eu}
]{\inits{AL}\fnm{Antonio} \snm{Longa}}
\author[
   addressref={aff6},
   email={lepri@fbk.eu}
]{\inits{BL}\fnm{Bruno} \snm{Lepri}}
\author[
   addressref={aff1},
   corref={aff1}, 
   email={luca.pappalardo@isti.cnr.it}
]{\inits{LP}\fnm{Luca} \snm{Pappalardo}}


\address[id=aff1]{
  \orgname{Institute of Information Science and Technologies, National Research Council (ISTI-CNR)}, 
  \city{Pisa},                              
  \cny{Italy}                                    
}

\address[id=aff2]{
  \orgname{IMT School for Advanced Studies}, 
  \city{Lucca},                              
  \cny{Italy}                                    
}

\address[id=aff3]{
  \orgname{University of Pisa}, 
  \city{Pisa},                              
  \cny{Italy}                                    
}
\address[id=aff4]{%
  \orgname{Free University of Bolzano},
  \city{Bolzano},
  \cny{Italy}
}

\address[id=aff5]{
  \orgname{University of Trento}, 
  \city{Trento},                              
  \cny{Italy}                                    
}
\address[id=aff6]{%
  \orgname{Fondazione Bruno Kessler},
  \city{Trento},
  \cny{Italy}
}



\end{fmbox}


\begin{abstractbox}

\begin{abstract} 

The increasingly crucial role of human displacements in complex societal phenomena, such as traffic congestion, segregation, and the diffusion of epidemics, is attracting the interest of scientists from several disciplines.
In this article, we address mobility network generation, i.e., generating a city's entire mobility network, a weighted directed graph in which nodes are geographic locations and weighted edges represent people's movements between those locations, thus describing the entire mobility set flows within a city.
Our solution is MoGAN, a model based on Generative Adversarial Networks (GANs) to generate realistic mobility networks.
We conduct extensive experiments on public datasets of bike and taxi rides to show that MoGAN outperforms the classical Gravity and Radiation models regarding the realism of the generated networks.
Our model can be used for data augmentation and performing simulations and what-if analysis.

\end{abstract}




\end{abstractbox}
%

\end{frontmatter}



\section{Introduction}

The increasing complexity of urban environments \cite{batty2013new, andrienko2020sobigdata} and the crucial role played by human displacements in the diffusion of epidemics, not least the COVID-19 pandemic \cite{lucchini2021living, pepe2020covid, lai2019measuring, Ruktanonchai1465, kraemer2020effect, oliver2020mobile}, have created a great deal of interest around the study of individual and collective human mobility \cite{luca2021survey, wang2019urban, barbosa2018human}.
The prevention of detrimental collective phenomena such as traffic congestion, air pollution, segregation, and epidemics spread, which is crucial to make our cities inclusive, safe, resilient, and sustainable \cite{le2015towards, kroll2019sustainable, assembly2015sustainable, bohm2022gross}, depends on how accurately we can predict and simulate people's movements within an urban environment.

In this regard, a particularly challenging task is generating realistic mobility flows, i.e., flows of people among a set of geographic locations given their demographic and geographic characteristics (e.g., population and distance) \cite{luca2021survey, simini2021deep, barbosa2018human, wang2019urban, masucci2013gravity}.
Traditionally, flow generation is addressed through the Gravity model \cite{carey1867principles, zipf1946p, lenormand2016systematic, barbosa2018human, erlander1990gravity, luca2022modeling}, the Radiation model  \cite{simini2012universal, barbosa2018human, wang2019urban}, and their variants \cite{simini2021deep, barbosa2018human, prieto2018gravity, yan2017universal}.
The Gravity model assumes that the number of travelers between two locations (flow) increases with the locations’ populations while decreasing with the distance between them. 
The Radiation model is a parameter-free model that only requires information about geographic locations (e.g., population) and their intervening opportunities. 
The Gravity and the Radiation models are designed to generate single flows between pairs of locations and are typically used to complete a network in which some mobility flows are missing.

In this paper, we address \emph{mobility network generation}, a variation of flow generation that consists in generating a city's entire mobility network. 
A mobility network is a weighted directed graph in which nodes are geographic locations and weighted edges represent people's movements between those locations, thus describing the entire set of mobility flows within a city.

Our solution to mobility network generation -- MoGAN (Mobility Generative Adversarial Network) -- is based on Generative Adversarial Networks (GANs) \cite{goodfellow2014generative}, deep learning architectures composed of a discriminator, which maximizes the probability to classify real and artificial mobility networks correctly, and a generator, which maximizes the probability to fool the discriminator producing artificial mobility networks classified by the discriminator as real. 
The choice of GANs is motivated by the fact that mobility networks can be represented as weighted adjacency matrices, similarly to how images are typically represented, and considering that GANs are tremendously effective in generating realistic images \cite{goodfellow2014generative, creswell2018generative, goodfellow2016nips, radford2016unsupervised}. 
While several papers show that GANs can generate individual mobility trajectories \cite{luca2021survey, liu2018trajgans, yin2018gans, kulkarni2018generative, ouyang2018non, huang2019variational, feng2020learning, moreira2013predicting} with a realism comparable to or better than mechanistic mobility models \cite{cornacchia2021modelling, pappalardo2017data, jiang2016timegeo, barbosa2018human}, to what extent GANs can generate realistic mobility flows has never been explored in the literature.

We train MoGAN on a set of real mobility networks and develop a tailored evaluation methodology to test the model's effectiveness in generating realistic mobility networks. 
We conduct extensive experiments on four public mobility datasets, describing flows of bikes and taxis in New York City and Chicago, US, to demonstrate that MoGAN generates synthetic mobility networks that are way more realistic than those generated by several baseline models, i.e., the Gravity, the Radiation, and the Random Weighted models.
Our results prove that our solution can synthesize aggregated movements within a city into a realistic generator, which can be used for data augmentation and performing simulations and what-if analysis.

\section{Mobility network generation}
\label{sec:res}

Mobility network generation consists of generating a realistic mobility network, i.e., a weighted directed graph in which nodes are locations and edges represent flows between those locations. 
The locations are defined by a discretization of the geographic space defined by a spatial tessellation, i.e., a covering of the bi-dimensional space using a countable number of geometric shapes called tiles, with no overlaps and no gaps \cite{luca2021survey}. 
In mobility networks, nodes are tiles of the spatial tessellation and edges flows of people among these tiles. 

Formally, we define a mobility network as a weighted directed graph  $\mathcal{G} = (V, E, w)$, where:
\begin{itemize}
    \item $V$ is the set of nodes, i.e., tiles of the spatial tessellation;
    \item $w: V \times V \mapsto \mathbb{N} $ is a function that assigns to each pair of nodes the number of people moving between the two nodes (mobility flow); 
    \item $ E = \{(x,y) | (x,y) \in V \times V  \land w(x,y) \neq 0 \} $ is the set of the weighted directed edges in the network.
\end{itemize}

A mobility network may contain self-loops (edges in which the origin and destination coincide), which describe movements of people within the same tile. 
Here, we represent a mobility network as a weighted adjacency matrix $\mathcal{A}_{n \times n}$ with $n = |V|$. 
Thus, an element $a_{i,j} \in \mathcal{A}$ represents the number of people moving from node \textit{i} to node \textit{j}, with $i, j \in V$. 

A generative model of mobility networks $M$ is any algorithm able to generate a set of $n$ synthetic mobility networks $\mathcal{X}_M = \{ \hat{\mathcal{G}}_1, \dots, \hat{\mathcal{G}}_n \}$, which describe the set of mobility flows on a given spatial tessellation. 
The realism of $M$ is evaluated with respect to:
\begin{enumerate}
\item A set of network patterns $\mathcal{K} = \{ s_1, \dots, s_{m}\}$ that describe some statistical properties of mobility networks. 
A realistic set $\mathcal{T}_M$ of synthetic mobility networks is expected to reproduce as many of these mobility patterns as possible.

\item A set $\mathcal{X} = \{ \mathcal{G}_1, \dots, \mathcal{G}_n \}$ of real mobility networks that describe real flows on the same spatial tessellation. 
Typically, a portion $\mathcal{X}_{\mbox{\footnotesize train}} \subset \mathcal{X}$ is used to train $M$ or to fit its parameters. The remaining part $\mathcal{X}_{\mbox{\footnotesize test}} \subset \mathcal{X}$ is used to compute the set $\mathcal{K}$ of patterns, which are compared with the patterns computed on $\mathcal{X}_M$.

\item A function $D$ that computes the dissimilarity between two distributions. 
Specifically, for each measure in $f \in \mathcal{K}$, $D(P_{(f, \mathcal{X}_M)}||P_{(f, \mathcal{X}_{\mbox{\footnotesize test}})})$ indicates the dissimilarity between $P_{(f, \mathcal{X}_M)}$, the distribution of the measures computed on the synthetic mobility networks in $\mathcal{X}_M$, and $P_{(f, \mathcal{X}_{\mbox{\footnotesize test}})}$, the distribution of the measures computed on the mobility networks in $\mathcal{X}_{\mbox{\footnotesize test}}$.
The lower $D(P_{(f, \mathcal{X}_M)}||P_{(f, \mathcal{X}_{\mbox{\footnotesize test}})})$, the more realistic model $M$ is with respect to $f$ and $\mathcal{X}_{\mbox{\footnotesize test}}$.
\end{enumerate}

\section{MoGAN: A Mobility Generative Adversarial Network}
To solve the problem of mobility network generation, we design MoGAN (Mobility Generative Adversarial Network), a deep learning architecture based on Deep Convolutional Generative Adversarial Networks (DCGANs) \cite{radford2016unsupervised}. 
MoGAN consists of a generator $G$, which learns how to produce new synthetic mobility networks, and a discriminator $D$, which has the task of distinguishing between real and fake (artificial) mobility networks. 
$G$ and $D$ are trained in an adversarial manner: $D$ maximizes the probability to correctly classify real and fake mobility networks; $G$ maximizes the probability to fool $D$, i.e., to produce fake mobility networks classified by $D$ as real. 
Both $D$ and $G$ are Convolutional Neural Networks (CNNs), which are proven to be effective in capturing spatial patterns in the data \cite{luca2021survey}.

During the training phase, $G$ repeatedly takes a $1 \times 100$ noise vector as input and operates a series of transposed convolutions, which perform upsampling of the input vector to generate a $64 \times 64$ adjacency matrix representing a mobility network.
Then, $D$ takes a set of real and generated $64 \times 64$ matrices as input and performs a binary classification task to classify these matrices as real or fake. 
The above process is repeated for a certain number of epochs and stopped when some criteria are met (see Supplementary Note 1).

MoGAN leverages the architecture of DCGAN \cite{radford2016unsupervised} and, as highlighted above, this implies that the shape of adjacency matrices must be $64\times64$. 
MoGAN could easily be extended to geographic areas with less than 64 zones, for example testing MoGAN ability of working with 0-padded mobility networks. On the other hand, working with more than 64 zones would necessarily require some form of aggregation of the zones, or a totally different GAN structure.

Once MoGAN is trained, $G$ can be used to generate as many mobility networks as desired. A visual representation of the networks generated during the training phase can be found in Supplementary Note 2.
Figure \ref{fig:ConvMNGAN} schematizes and describes MoGAN's architecture.
Further details on MoGAN's architecture and training can be found in Supplementary Note 1.

\begin{figure}[h!]
\centering
    \includegraphics[width=.92\textwidth]{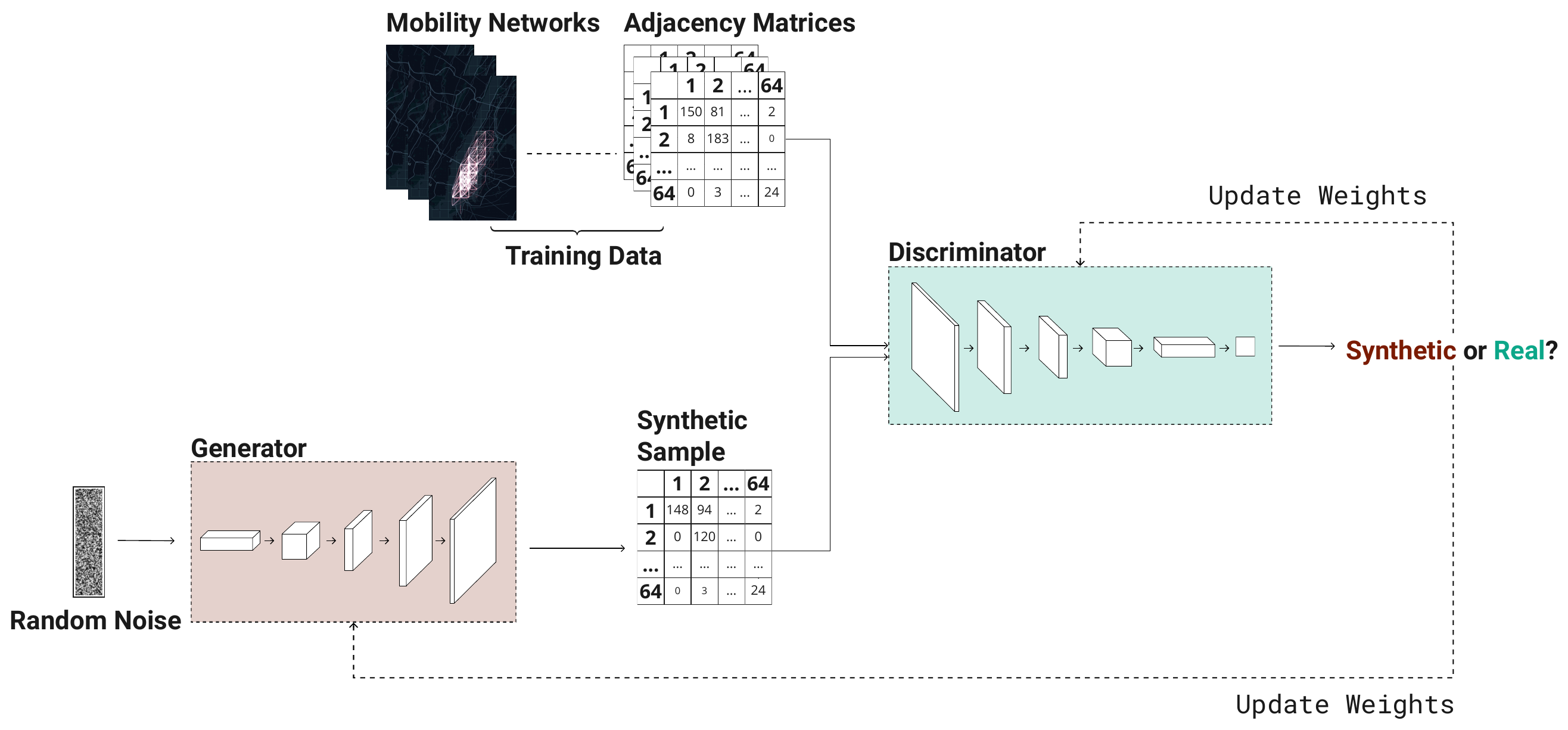}
  \caption{\csentence{Architecture of MoGAN.}
{The generator (a Convolutional Neural Network or CNN) performs transposed convolution operations that upsample the input random noise vector, transforming it into a $64\times 64$ adjacency matrix representing a mobility network. 
The discriminator (a CNN) takes as input both the generated mobility networks and the real ones from the training set and performs a series of convolutional operations that end up with a probability, for each sample, to be fake or real. 
Both the discriminator's and generator's weights are then backpropagated.}}
    \label{fig:ConvMNGAN}
\end{figure}

\section{Baseline models}
\label{sec:base}
We compare MoGAN with the Gravity and the Radiation models, two classical approaches for mobility flows' generation \cite{barbosa2018human, luca2021survey, simini2021deep, simini2012universal}, using the implementations provided in library scikit-mobility \cite{pappalardo2019scikitmobility}.

The singly-constrained Gravity  model~\cite{carey1867principles, zipf1946p, lenormand2016systematic, barbosa2018human} prescribes that the expected flow, $\bar{y}$, between an origin location $l_i$ and a destination location $l_j$ is generated according to the following equation:
\begin{equation}
\label{eq:gravity}
\bar{y}(l_i,l_j) = O_i p_{ij} = O_i \frac{m_j^{\beta_1}f(r_{ij})}{\sum_k m_k^{\beta_1} f(r_{ik})}
\end{equation}
where $O_i$ is the number of people leaving location $l_i$, $m_j$ is the population of location $l_j$ (estimated as $O_j$), 
$p_{ij}$ is the probability to observe a trip (unit flow) from location $l_i$ to location $l_j$, $\beta_1$ is a parameter and $f(r_{ij})$ is the deterrence function, which is a function of the distance $r_{ij}$ between two locations. 
We model the deterrence function as a power-law function, $f(r) = r^{\alpha}$, where $\alpha$ is another parameter.
These parameters can be fitted from a  subset of available flows. We report the value of $\alpha$ and $\beta_1$ resulting from the fitting of the model in Supplementary Note 3.

The Radiation model \cite{simini2012universal, barbosa2018human} is a parameter-free model that aims to generate flows between locations given their characteristics (e.g., population) and the intervening opportunities among them. 
The choice of the destination consists of two steps: \emph{(i)}
we assign a fitness $z$ to each location opportunity sampled from a distribution $p(z)$ that represents the quality of the opportunity for each travel; \emph{(ii)} the traveler ranks the opportunities according to their distance from the origin location and chooses the nearest location with a fitness higher than a certain threshold.
As a result, the mean flow between two locations $l_i$ and $l_j$ is calculated as:
\begin{equation}
\label{eq:radiation}
\bar{y}(l_i, l_j) =O_i  \frac{1}{1-\frac{m_i}{M}} \frac{m_i  m_j}{(m_i + s_{ij}) (m_i + m_j +s_{ij})}
\end{equation}
where $O_i$ is the number of people leaving location $l_i$, $m_i$ and $m_j$ are the opportunities in $l_i$ and $l_j$, $M$ is the sum of all the opportunities, and $s_{ij}$ is the number of opportunities in a circle of radius $r_{ij}$.

Note that the Gravity and the Radiation models do not solve mobility network generation directly.
While MoGAN, once trained, can generate an entire mobility network, the Gravity and the Radiation models are designed to generate single flows between pairs of locations.
To generate a mobility network using the Gravity and the Radiation models, we proceed as follows: \emph{(i)} we take a real mobility network; \emph{(ii)} for each node, we compute its relevance $m_i$ and total outflow $O_i$; and \emph{(iii)} we use $m_i$ and $O_i$ in Equations \ref{eq:gravity} and \ref{eq:radiation}.
For the Gravity model, we also fit parameters $\beta_1$ and $\beta_2$ from the real mobility network assuming a power-law deterrence function. 
For both the Gravity and Radiation models, we use the implementations available in the library scikit-mobility \cite{pappalardo2019scikitmobility}, which provides methods to fit parameters and generate flows from locations' relevance and outflow. 

For a further analysis, we compare MoGAN with a
Random Weighted (RW) model that creates a mobility network where the weight of each edge is randomly chosen from the distribution of weights for that edge in the training set. 
In other words, given an edge $e = (i,j)$ connecting node $i$ to node $j$ in the mobility network, the edge weight $\hat{w}(e)$ is a number picked at random from $\{ w_1(e), w_2(e), \dots w_n(e) \}$, i.e., the distribution of the weights of $e$ in the training set.

In terms of computational time required to generate a new mobility network, MoGAN is way faster (< 1 second) than the Gravity model (about one minute) and the Random Weighted model (10-20 seconds). 
However, MoGAN needs a training phase that requires from 1 up to 3 hours depending on the dataset.  
In out experiments, we train MoGAN on a server with a GPU Tesla P100 with 16GB of VRAM, 13GB of RAM and a 2-core Intel Xeon CPU.

\section{Experimental setup}
\subsection{Datasets}
We use four real-world public datasets, which describe trips with taxis and bikes in New York City and Chicago during 2018 and 2019 (730 days). 
Two datasets contain daily information regarding the use of bike-sharing services: the City Bike Dataset for New York City \cite{citi} and the Divvy Bike Dataset for Chicago \cite{divvy}.
Each record describes the coordinates of each ride's starting and ending station, and the starting and ending times. 
We remove trips with a duration lower than 60 seconds because they could be false starts or users trying to re-dock a bike to ensure it is secure \cite{citi, divvy}.
We also use two datasets containing daily information about the movements of taxis: the New York City taxi dataset \cite{taxi_nyc} and the Chicago taxi dataset \cite{taxi_chi}.
A record describes each ride's starting and ending location and the starting and ending times. 
Both datasets are already preprocessed to remove dummy and noisy rides. 
In the Chicago taxi dataset, we know the GPS points corresponding to the starting and ending points of each taxi trajectory. 
In the New York City taxi dataset, we only know the trajectories' starting and ending zones, i.e., administrative areas in New York City.
We use an administrative area's centroid as a taxi ride's reference starting or ending point.
We select the island of Manhattan for New York City and the central districts for Chicago (see Supplementary Figure S3) and split the selected zones into 64 equally-sized squared tiles (1840 meters per side for New York City, 1405 meters per side for Chicago).
For each dataset, we count the daily number of taxis or bikes moving between each pair of tiles to obtain an origin-destination matrix representing the daily mobility network. 
We obtain, for each dataset, a representation of the daily flows in the city, which is divided into 64 equally spaced tiles. 
The mobility networks represent the flow of people moving, daily, across these zones.
We remind that, since MoGAN is based on DCGANs which are designed to work with images (matrices) of size $64 \times 64$, we are constrained to tessellate the city into 64 equally-sized tiles. 
This means that different cities have different tile size, depending on the city size.

\begin{figure}[htb!]
\begin{center}
\includegraphics[width=.95\textwidth]{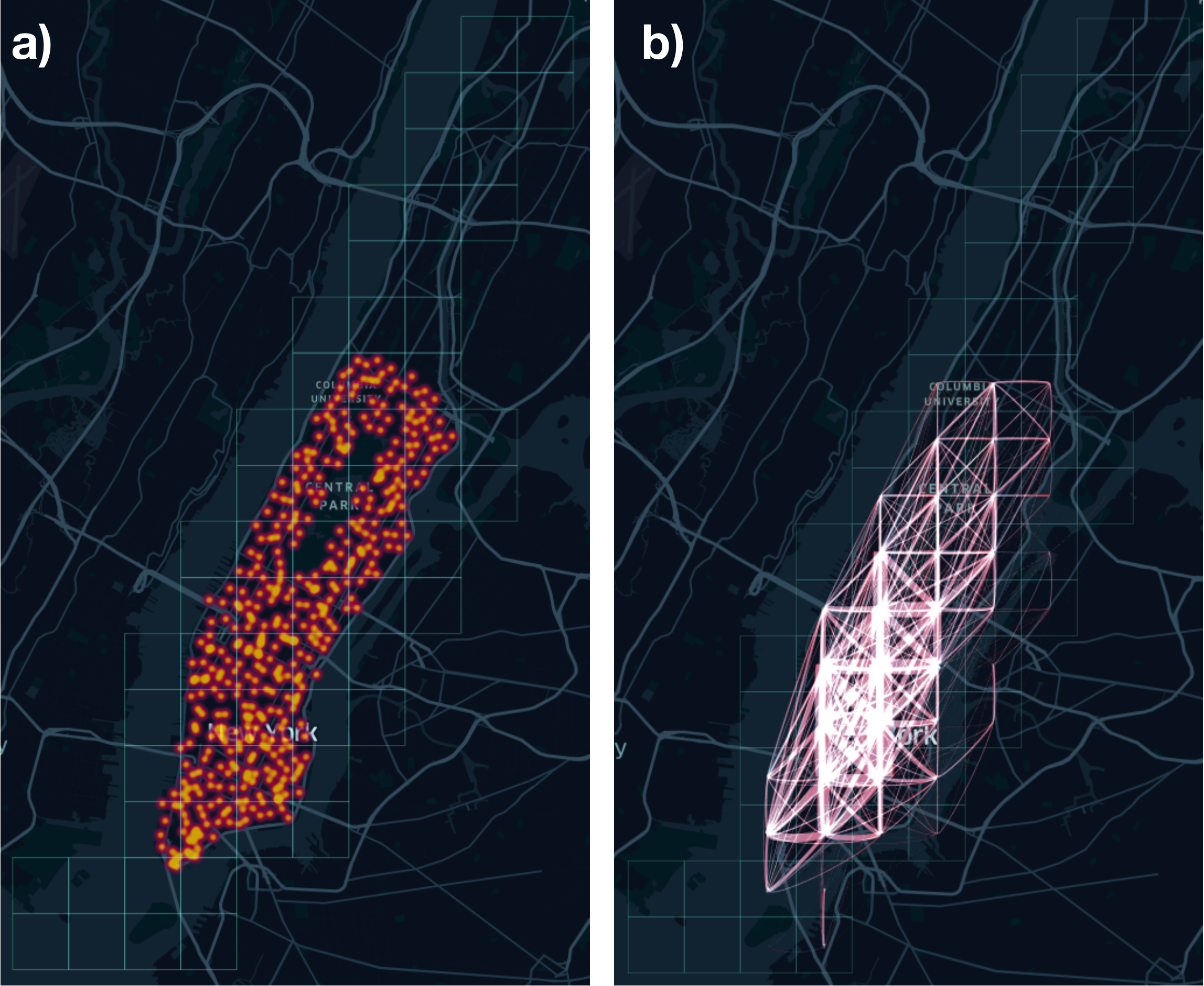}
  \caption{\csentence{Examples of a real mobility network.}
      {(a) Position of bike stations in Manhattan.
    (b) A daily mobility network in Manhattan, where the size of each edge is proportional to the flow they represent.}
    }
    \label{fig:final_imgs}
\end{center}
\end{figure}

We compute the relevance of each location (tile), which is needed for generating flows in the Gravity and the Radiation models, as the total number of daily drop-offs in that location.
Table \ref{tab:data} shows some statistics about the datasets used in our study.
As an example, Figure \ref{fig:final_imgs} visualizes where bike stations concentrate and a mobility network representing daily flows in Manhattan, New York City.

\begin{table}[htb!]
\centering
\resizebox{.8\textwidth}{!}{%
\begin{tabular}{|c|c|c|c|c|} 
\hline
\textbf{dataset} & \textbf{rides} & \textbf{locations} & \textbf{\#bikes/taxis}  \\ 
\hline
CHI bikes \cite{divvy}       & 3,505,03          & 198                  & 6293\\ 
\hline
NYC bikes \cite{citi}        & 29,294,326        & 509                 & 19514           \\ 
\hline
CHI taxis \cite{taxi_chi}       & 11,050,936       & 96                  & 5668            \\ 
\hline
NYC taxis \cite{taxi_nyc}       & 157,485,483     & 68                  & N.D.            \\
\hline
\end{tabular}
}
\caption{Statistics of the four datasets used in our study. 
For each dataset, we provide the link to download it, the number of rides, the different locations, and the number of bikes/taxis. 
CHI = Chicago, NYC = New York City.
For the NYC taxi dataset, taxi identifiers are not available and we do not know the total number of taxis. 
All datasets refer to trips in 2018 and 2019.}
\label{tab:data}
\end{table}

\subsection{Validation}
\label{sec:val}
We develop a tailored approach to evaluate the realism of the mobility networks generated by MoGAN. 
For each dataset, we construct a mobility network for each day obtaining 730 real mobility networks in total.
We split the 730 networks into a training set (584 networks) and a test set (146 networks).
We train MoGAN on the training set and generate 146 synthetic mobility networks (synthetic set). 
We evaluate the model's realism computing the difference between each network in the synthetic set and each network in the test set, so obtaining $146\times 146=21,316$ values.
If the generated mobility networks are realistic, they should differ from the real networks to the same extent real networks differ between themselves.
To stress this aspect, we create a set of 146 mobility networks (mixed set), in which half of them are chosen uniformly at random from the test set, and the other half is chosen uniformly at random from the synthetic set.
We then compute the pairwise difference between any possible pair of mobility networks in the mixed set.

The idea behind this validation methodology is that we need to verify whether MoGAN is capable to reproduce the variability of mobility networks in the training set. 
If the distribution of differences among the networks in the synthetic set is similar to that of networks in the test set, MoGAN can approximate well the variability of mobility networks in the training set. 
The use of the mixed set further tests MoGAN’s ability to reproduce the variability in the training set: by verifying that the distribution of differences between networks in the synthetic set and those in the test set is similar to the distribution of network distances within the test set and the synthetic set separately, we argue that MoGAN can reproduce the variability of networks in the training set.

A crucial aspect is how to compute the difference between two mobility networks, considering that directed weighted networks are hard to compare, even in the case of known-node correspondence (i.e., networks with the same nodes but different edges) \cite{tantardini2019comparing}. 
We compute this difference in two ways. 

The first one consists of computing an error metric between two networks' adjacency matrices. 
In our experiments, we try three error metrics: \emph{(i)} Normalized Root Mean Square Error (NRMSE), \emph{(ii)} Common Part of Commuters (CPC), and \emph{(iii)} Cut Distance (CD). 
The Root Mean Square Error (RMSE) \cite{luca2021survey, simini2021deep} is defined as: $$\mathrm{RMSE}(A,B) = \sqrt{\frac{1}{n} \sum_{i,j=1}^n (a_{ij} - b_{ij})^2 } $$

\noindent where $a_{ij}$ and $b_{ij}$ are the elements (flows) in position $(i,j)$ in the two networks' adjacency matrices of $A$ and $B$ and $n$ is the number of elements of the matrices ($64\times64$). 
Note that RMSE is substantially equivalent to the Frobenious norm (see Supplementary Note 5). The NRMSE is a min-max normalization of the RMSE, defined as: $$ \mathrm{NRMSE} = \frac{RMSE(A,B)}{max(A,B) - min(A,B)}$$

The Common Part of Commuters (CPC), also known as Sørensen-Dice index \cite{barbosa2018human, simini2021deep, LENORMAND2016158, luca2021survey}, a well-established measure to compute the similarity between real and generated matrices, is defined as: 
    $$\mathrm{CPC}(A,B) = \frac{2\sum_{i,j=1}^n min(a_{ij}, b_{ij})}   {\sum_{i,j=1}^n a_{ij} + \sum_{i,j=1}^n b_{ij} }$$
    
CPC is a widely used metric in human mobility studies \cite{luca2021survey, lenormand2016systematic} and it ranges between 0 and 1. A CPC of 1 indicates a perfect match between the generated flows and the ground truth. On the other hand, 0 highlights a bad performance with no overlap. In other terms, CPC can be seen and interpreted as a metric of accuracy.

The Cut Distance (CD) \cite{liu2018cut} is based on the notion of cut weight, widely used in network theory \cite{tantardini2019comparing}, and measures how much a network is bipartite. 
The cut norm $||A||_C$ of a real matrix $A = (a_{ij} )$, $i\in R$, $j \in S$ with a set of rows indexed by $R$ and a set of columns indexed by $S$, is the maximum over all $I \subset R$, $J \subset S$ of the quantity $|\sum_{i\in I, j\in J} a_{ij} |$. 
The Cut Distance (CD) between two adjacency matrices $A$ and $B$ is the cut norm of their difference:

$$\mathrm{CD}(A,B) = \max_{S \in V}    \frac{1}{|V|} \left| e_{A}(S, S^C) -  e_{B}(S, S^C)   \right|$$

\noindent with $V$ being the number of nodes (64, in our case), $e_G(S,T) = \sum_{i \in S, j\in T} w_{ij} $ is the cut weight of adjacency matrix $G$ with weights $w_{ij}$, i.e., the sum of the weights of the edges that starts in $S$ and ends in $T$ and $S^{C} = V \setminus S$.
\cite{alon2004approximating}.
Maximizing this quantity is a computationally heavy problem, so we use the Semidefinite Program (SDP) approximation proposed by Chan and Sun \cite{o2008optimal}. For calculating CD, we use the python implementation available in the library cutnorm \cite{cutnorm}.

The second approach to computing the difference between two mobility networks consists of comparing their distributions of edge weights and weight-distances.
Edge weights indicate the values (flows) of the adjacency matrices describing the two mobility networks.
Weight-distances indicate the combination of an edge's weight (flow) and the distance between the two nodes composing the edge. 
We compute the weighted-distance adjacency matrix of a mobility network as $\hat{A} = A / (d+\epsilon)$, where $A$ is the network's weighted adjacency matrix, $d$ is the distance matrix having the same dimension and node ordering of $A$ and representing the geographic distances between all pair of nodes.\footnote{The geographic distance between two nodes is calculated as the distance between the centroids of the tile that represents that node.} 
We add the residual term $\epsilon = 0.01$ to the denominator just to avoid dividing by zero only for elements on the diagonal of the adjacency matrices. 
Given two mobility networks, the more similar their distribution of edge weights or weight-distances are, the more similar the two mobility networks are.
We measure the similarity between two distributions using the Jensen-Shannon divergence \cite{fuglede2004jensen, pappalardo2017data}:
$$ \mathrm{JS}(P || Q) = \frac{1}{2} \mathrm{KL}(P || M) + \frac{1}{2} \mathrm{KL}(Q || M) $$
where $P$ and $Q$ are two density distributions, 
$M = \frac{1}{2} (P + Q)$, and $KL$ is the Kullback–Leibler divergence (KL) \cite{kullback1997information, van2014renyi}, defined as:
$$\mathrm{KL}(P||Q) = \mathlarger{\sum}_{x \in X} P(x) log \left(\frac{P(x)}{Q(x)} \right)$$

An alternative to the usage of divergence metrics may consist in using kernels to measure similarities between graphs \cite{vishwanathan2010,nikolentzos2021}.  However, while kernel methods compare networks' representation in a latent space, in this paper we aim to capture the mobility network’s topological macro-scale features (e.g., degree distribution, clustering coefficient).
For the sake of completeness, in Supplementary Note 10, we provide a comparison between topological properties of the generated and real mobility networks such as the clustering coefficient and the weighted degree distribution.

\section{Results}

\begin{figure}[htb!]
\centering
    \includegraphics[width=.95\textwidth]{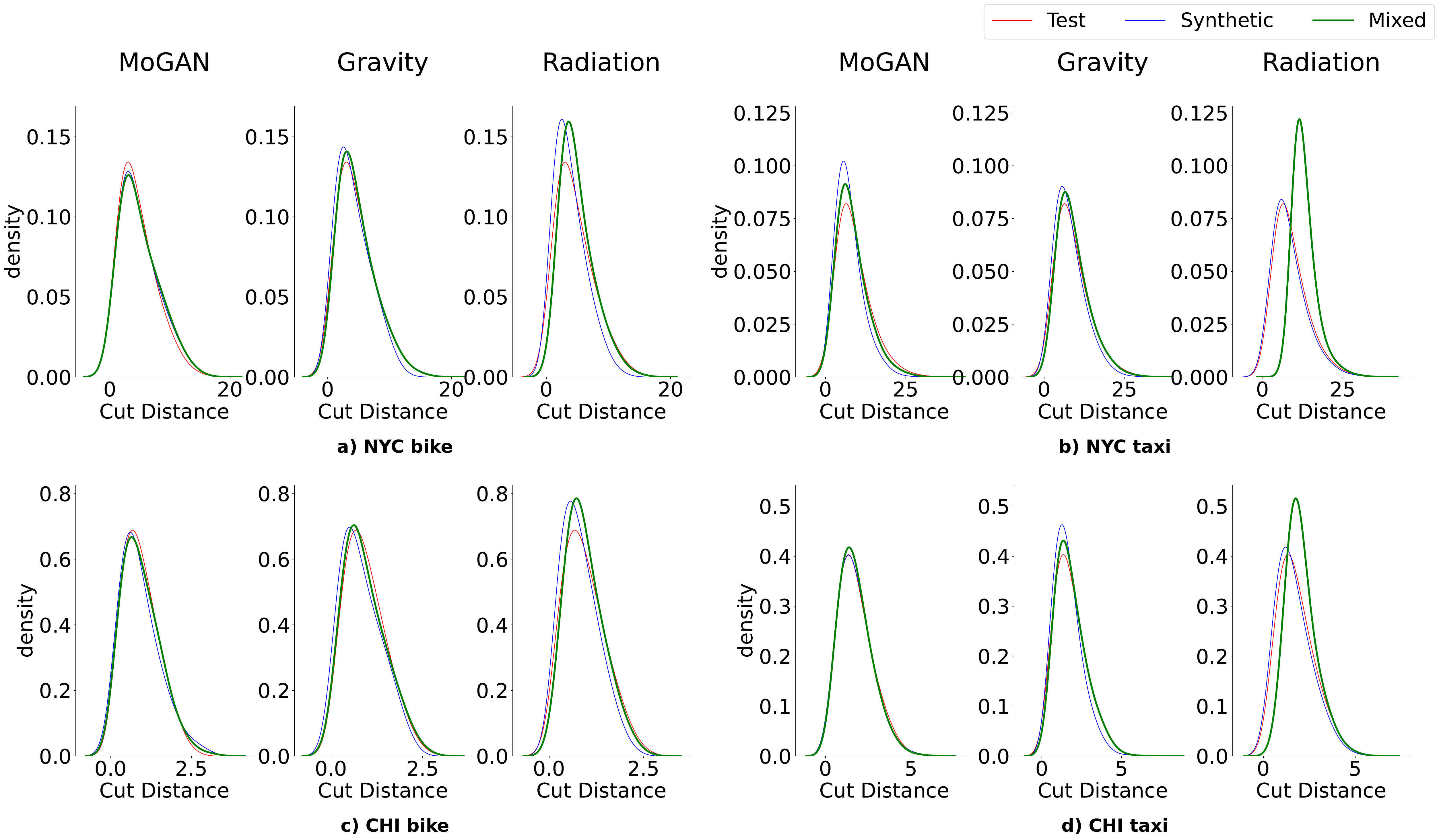}
  \caption{\csentence{Results for the Cut Distance. }
      {Distributions of the pairwise cut distances between mobility networks in the test set (red), synthetic set (blue), and mixed set (green), for the four datasets. 
      For each dataset, we compare the overlap among the distributions of MoGAN and the two baselines (Gravity and Radiation). 
      The Radiation model's mixed and synthetic sets distributions significantly differ from the test set for all datasets. 
      In contrast, the Gravity model clearly outperforms the Radiation model for all datasets.}}
    \label{fig:cut}
\end{figure}

\begin{figure}[htb!]
\centering
    \includegraphics[width=.95\textwidth]{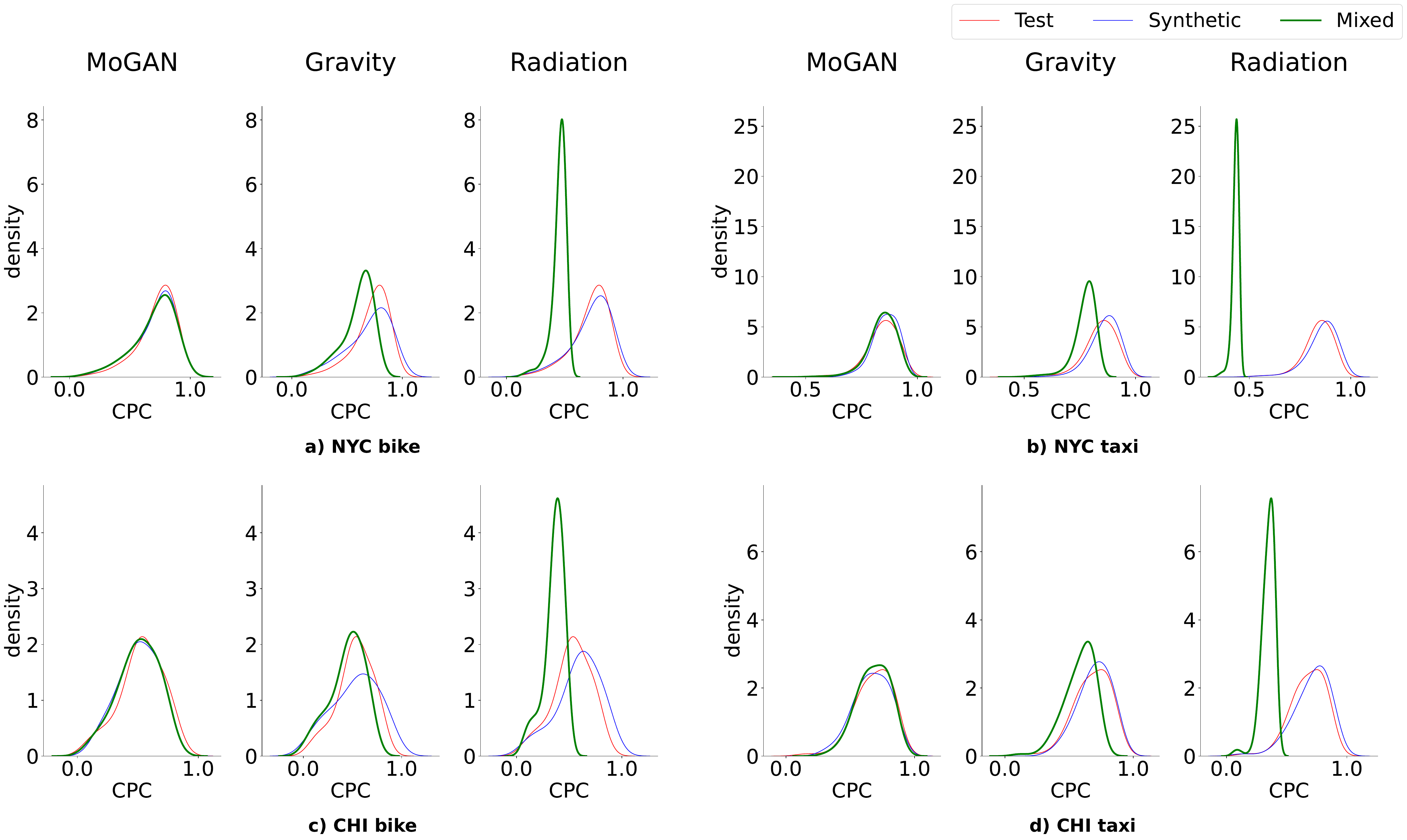}
  \caption{\csentence{Results for the CPC. }
      {Distributions of the pairwise CPC distances between mobility networks in the test set (red), synthetic set (blue), and mixed set (green), for the four datasets. 
      For each dataset, we compare the overlap of the distributions of MoGAN and the two baselines (Gravity and Radiation). For both the Gravity model and the Radiation model, the three distributions are significantly different, especially for the latter.}}
    \label{fig:cpc}
\end{figure}

Figure \ref{fig:cut} shows the distribution of the Cut Distance (CD) in the four datasets' test (red), synthetic (blue), and mixed sets (green) for MoGAN (left), the Gravity model (center), and the Radiation model (right).  
MoGAN's CD distributions overlap almost entirely in all four datasets, meaning that MoGAN generates mobility networks that are indistinguishable from real ones and way more realistic than those generated by the baselines (except in two cases, see Supplementary Note 6).
Similar results hold for the other metrics: MoGAN typically outperforms the baselines regarding CPC (Figure \ref{fig:cpc}) and RMSE (Supplementary Note 7). 
Table \ref{tab:res_cpc} shows, for each model, the JS-divergence between \emph{(i)} the CPC distribution of the mixed and test sets and \emph{(ii)} the CPC distribution of the synthetic and test sets. 

To compute the improvement in performance of MoGAN with respect to the baseline models, for each metric, each set and each baseline, we define the quantity:
$$\Delta = - \left(  \frac{JS^{(\mbox{\small MoGAN})} - JS^{(baseline)} }    {JS^{(baseline)}} \right)  \times 100$$
where $JS^{(\mbox{\small MoGAN})}$ is the $JS$ divergence between the set (synthetic or mixed) of networks generated by MoGAN and the test set, while $JS^{(baseline)}$ is the JS divergence between the set (synthetic or mixed) of networks generated by the baselines (Gravity, Radiation or Random Weighted) and the test set.

Table \ref{tab:res_cpc} shows that, according to the CPC, MoGAN outperforms the Gravity and Radiation models on all datasets, with a relative improvement of up to 86\% on the Gravity model and 91\% on the Radiation model over the mixed set, and a relative improvement of up to 49\% on the Gravity model and 37\% on the Radiation model over the synthetic set.
We report the results of the comparison with the Gravity and Radiation models for RMSE, CD, weights distribution and weight-distances distribution in Supplementary Notes 7, 8 and 9.

\begin{table}[htb!]
\centering
\resizebox{\textwidth}{!}{%
\begin{tabular}{|c||c|c||c|c||c|c|||c|c|c|c|}
\multicolumn{1}{l}{} & \multicolumn{2}{c}{\textbf{MoGAN}} & \multicolumn{2}{c}{Gravity} & \multicolumn{2}{c}{Radiation} & \multicolumn{4}{c}{Rel. Improvement}                       \\ 
\hline
data                 & $JS_m$ & $JS_s$                    & $JS_m$ & $JS_s$             & $JS_m$ & $JS_s$               & $\Delta_{m,G}$ & $\Delta_{s,G}$ & $\Delta_{m,R}$ & $\Delta_{s,R}$  \\ 
\hline
$NYC_{bike}$           & \textbf{0.06}   & \textbf{0.08}                      & 0.46   & 0.15               & 0.72   & 0.12                 & 86\%         & 49\%         & 91\%         & 37\%          \\ 
\hline
$NYC_{taxi}$           & \textbf{0.09}   & \textbf{0.11}                     & 0.53   & 0.14               & 0.83   & 0.15                 & 83\%         & 22\%         & 89\%         & 29\%          \\ 
\hline
$CHI_{bike}$           & \textbf{0.14}   & \textbf{0.16}                      & 0.29   & 0.25               & 0.56   & 0.26                 & 51\%         & 35\%         & 75\%         & 38\%          \\ 
\hline
$CHI_{taxi}$           & \textbf{0.08}   & \textbf{0.09}                      & 0.39   & 0.11               & 0.79   & 0.13                 & 80\%          & 21\%         & 90\%          & 30\%           \\
\hline
\end{tabular}
}
\caption{JS divergences of the distributions of the CPC scores.
For each model, we report the JS divergence between mixed set and test set (column $JS_{m}$) and the JS divergence between synthetic set and test set (column $JS_{s}$). 
The last four $\Delta_{x,Z}$-like columns represent the improvement of MoGAN compared to the Gravity model on the mixed and the synthetic sets (columns $\Delta_{m,G}$ and $\Delta_{s,G}$) and the improvement of MoGAN compared to the Radiation model on the mixed and synthetic sets (columns $\Delta_{m,R}$ and $\Delta_{s,R}$). }
\label{tab:res_cpc}
\end{table}

\begin{figure}[htb!]
\centering
    \includegraphics[width=.95\textwidth]{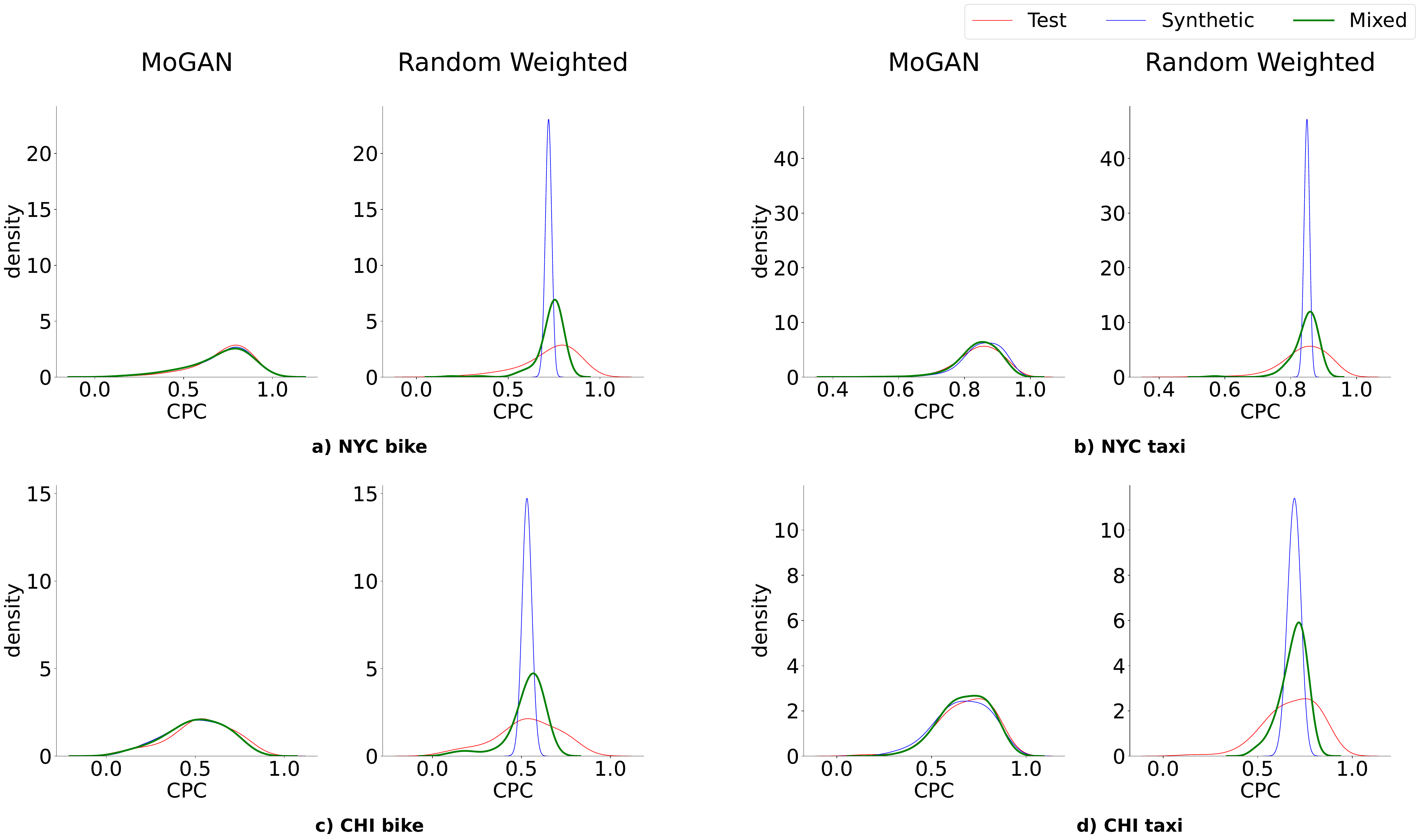}
  \caption{\csentence{Results for the CPC with Random Weighted Model.}
      {Distributions of the pairwise CPC distances between mobility networks in the test set (red), synthetic set (blue), and mixed set (green), for the four datasets. 
      For each dataset, we compare the overlap of the distributions of MoGAN and the Random Weighted model. 
      MoGAN distributions are perfectly overlapping, while the Random Weighted ones show significant differences.}}
    \label{fig:ran_cpc}
\end{figure}

MoGAN also outperforms the Random Weighted model for all proposed metrics. 
Figure \ref{fig:ran_cpc} compares the performance of MoGAN and the Random Weighted model according to CPC. 
For each dataset, MoGAN's test, synthetic and mixed set distributions are more overlapping than the ones of the Random Weighted model. We report the results of the comparison with Random Weighted model for RMSE, CD, weights distribution and weight-distances distribution in Supplementary Notes 11-14. 
Table \ref{tab:ran_res_cpc} shows that, according to CPC, MoGAN outperforms the Random Weighted model for all datasets.

\begin{table}[htb!]
\centering
\resizebox{.8\textwidth}{!}{%
\begin{tabular}{|c||c|c||c|c|||c|c|}
\multicolumn{1}{l}{} & \multicolumn{2}{c}{\textbf{MoGAN}}          & \multicolumn{2}{c}{Random Weighted}         & \multicolumn{2}{c}{Rel. Improvement}         \\ 
\hline
data                 & $JS_m$               & $JS_s$               & $JS_m$               & $JS_s$               & $\Delta_m,RW$        & $\Delta_s,RW$         \\ 
\hline
$NYC_{bike}$         & \textbf{0.06}                 & \textbf{0.08}                 & 0.45                 & 0.63                 & 86\%                 & 88\%                  \\ 
\hline
$NYC_{taxi}$         & \textbf{0.09}                 & \textbf{0.11}                 & 0.37                 & 0.59                 & 76\%                 & 82\%                  \\ 
\hline
$CHI_{bike}$         & \textbf{0.14}                 & \textbf{0.16}                 & 0.4                  & 0.56                 & 64\%                 & 71\%                  \\ 
\hline
$CHI_{taxi}$         & \textbf{0.08}                 & \textbf{0.09}                 & 0.37                 & 0.55                 & 79\%                 & 84\%                  \\ 
\hline
\multicolumn{1}{l}{} & \multicolumn{1}{l}{} & \multicolumn{1}{l}{} & \multicolumn{1}{l}{} & \multicolumn{1}{l}{} & \multicolumn{1}{l}{} & \multicolumn{1}{l}{}  \\
\multicolumn{1}{l}{} & \multicolumn{1}{l}{} & \multicolumn{1}{l}{} & \multicolumn{1}{l}{} & \multicolumn{1}{l}{} & \multicolumn{1}{l}{} & \multicolumn{1}{l}{} 
\end{tabular}
}
\caption{JS divergences of the distributions of the CPC scores with the Random Weighted model.
For each model, we report the JS divergence between mixed set and test set (column $JS_{m}$) and the JS divergence between synthetic set and test set (column $JS_{s}$). 
The last two $\Delta_{x,Z}$-like columns represent the improvement of MoGAN compared to the Random Weighted model on the mixed and the synthetic sets (columns $\Delta_{m,RW}$ and $\Delta_{s,RW}$).} 
\label{tab:ran_res_cpc}
\end{table}

MoGAN's JS-divergences between the mixed and test sets and between the synthetic and test sets are the lowest for each dataset, meaning that our model produces the most overlapping distributions (see Table \ref{tab:res_cpc}). 
Our results also show that the difference (either in terms of CD, CPC, or RMSE) between a real network and a synthetic one is similar to the difference between two real networks or two synthetic networks. 
This means that MoGAN generates realistic mobility networks that are, to a certain extent, indistinguishable from real ones.

\begin{figure}[htb!]
\centering
    \includegraphics[width=.95\textwidth]{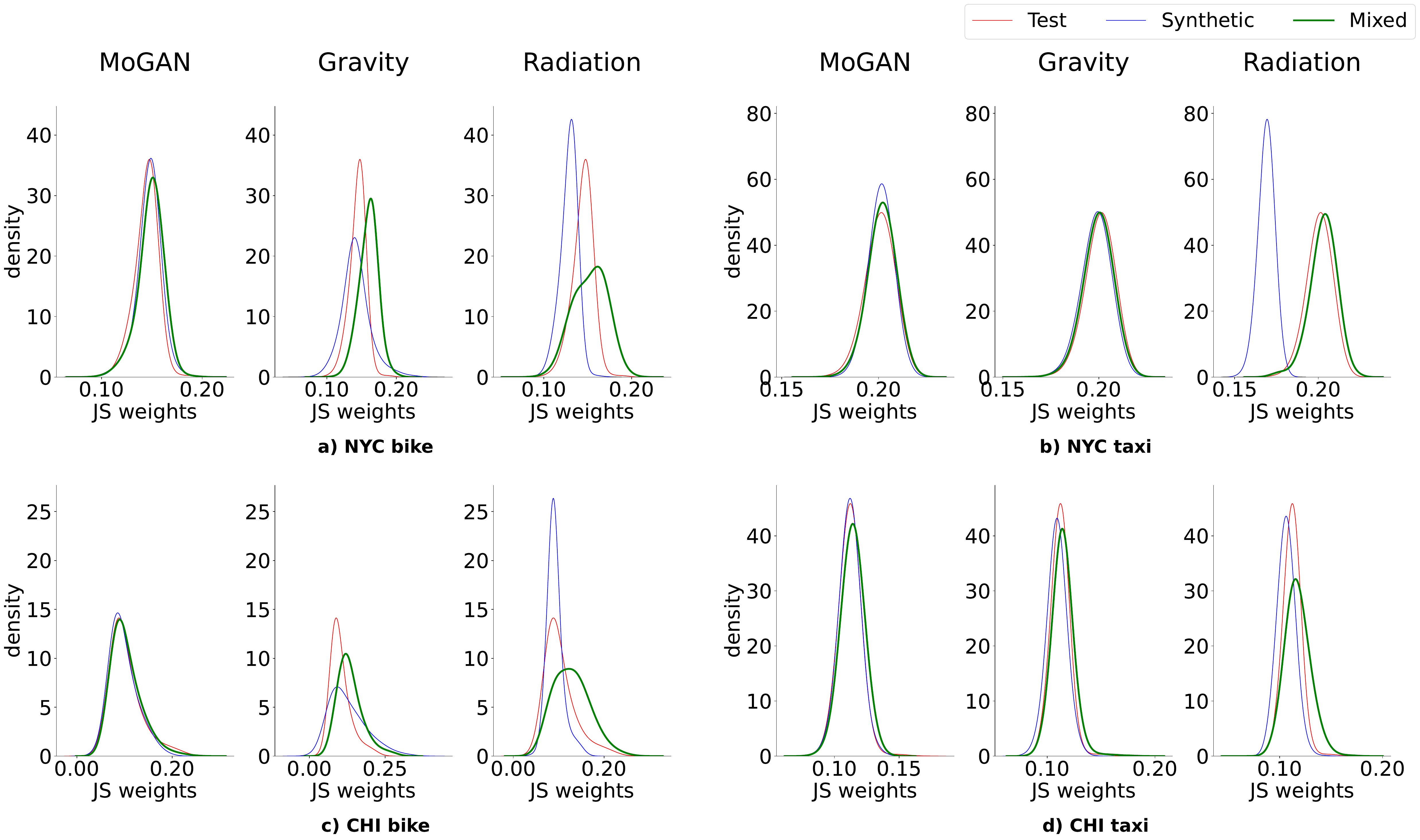}
  \caption{\csentence{Results for weights distribution. }
     {Distributions of the pairwise JS distance between the distribution of weights of the mobility networks in the test set (red), synthetic set (blue), and mixed set (green), for the four different datasets. 
     For each dataset, we compare the overlap of the distributions of MoGAN and the two baselines (Gravity and Radiation). 
     The Radiation model's mixed and synthetic sets distributions significantly differs from the test set for all datasets. 
     The situation is similar for the Gravity model performances. 
     MoGAN distributions are almost overlapping for all four datasets.}}
    \label{fig:weights}
\end{figure}

Figure \ref{fig:weights} shows the distributions of the pairwise similarities among the edge weights for the synthetic, mixed, and test sets built over the four datasets. 
For each dataset, we report the performances of MoGAN, the Gravity model, and the Radiation model.
Again, MoGAN significantly outperforms the baselines, except for two cases (mixed set of NYC and CHI taxi) in which the Gravity model and MoGAN achieve similar performance.
We find a similar result for the weight-distances (see Supplementary Note 7).

\begin{figure}[htb!]
\centering
    \includegraphics[width=.95\textwidth]{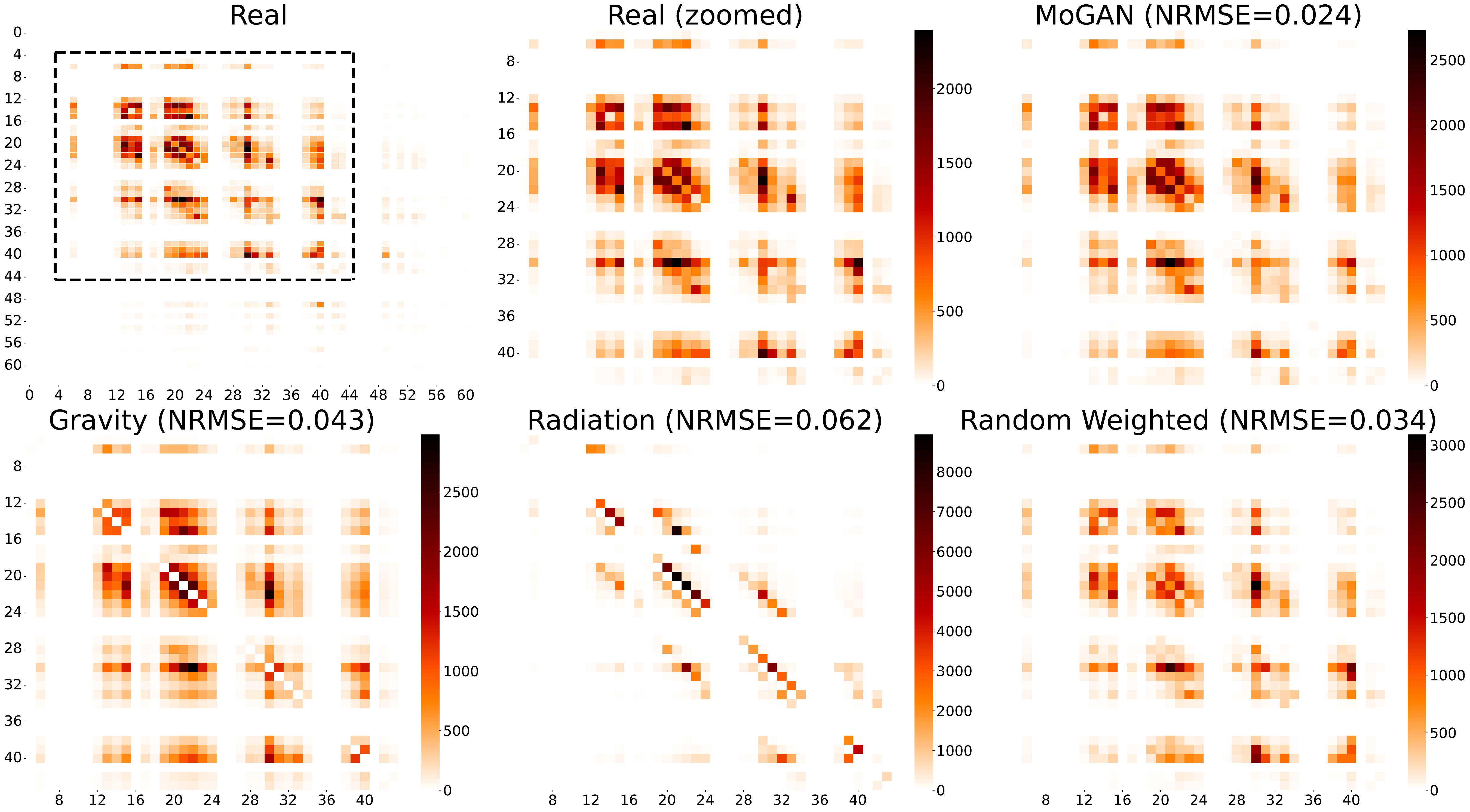}
  \caption{\csentence{Visual comparison of the adjacency matrices of the Mobility Networks. }
      {Visualization of the more dense part of the mobility networks of NYC Bikes having the maximum sum of flows observed in the Test Set (Real Zoomed) and of the Mobility Networks having the maximum sum of flows observed in the fake sets produced by all of the other models. Per each generated matrix, we reported the RMSE with respect to the Real matrix. In the top left panel, we show the full 64$\times$64 mobility network and highlight the most dense zones, on which we focus in the other plots of the figure.}}
    \label{fig:viz}
\end{figure}

In Figure \ref{fig:viz}, we compare a subset of the entries in the adjacency matrices representing the generated mobility networks with the adjacency matrix of a real mobility network. 
We compared only this part of the matrices for visualization reasons: the external part of them is, in fact, made up of 0 entries.
For each model, we visualize the generated mobility network with the maximum sum of flows. 
We observe that MoGAN's adjacency matrix is way more similar to the real one than the other models. 
The Gravity model produces an adjacency matrix that looks quite similar to the real one, but it lacks the self-loops (the elements on the diagonal). 
The Random Weighted model's matrix resembles the real one but the magnitude of flows differ in several parts of the network. 
The Radiation model's adjacency matrix is way different to the real one.

Figure \ref{fig:viz} shows that MoGAN is way better than the Gravity model at predicting flows between close tiles. In contrast, the two models reach a similar performance for flows regarding tiles that are very distant to each other. In Supplementary Note 15, we report the correlation between the error and the distance between flows' tiles for the BikeNYC dataset, for both MoGAN and the Gravity model.


\section{Conclusion}
\label{sec:concl}

This paper introduces MoGAN, a deep-learning-based model for generating realistic urban mobility networks.
Our results, conducted on four public datasets representing flows of bikes and taxis in New York City and Chicago, show that the realism of the networks generated by MoGAN outperforms those generated by classic models such as the Gravity and the Radiation models. 

Although MoGAN's performance is encouraging, it also has some limitations. 
Being based on DCGAN \cite{radford2016unsupervised}, MoGAN can generate $64\times64$ adjacency matrices, that is, mobility networks with 4096 locations. 
We plan to extend MoGAN's architecture to generate mobility networks with an arbitrary number of nodes as future improvements.
Other technical improvements may be the use of Graph Neural Networks (GNNs) \cite{scarselli2008graph}, which would better capture the network dependencies and include other location-related information (e.g., population or relevance), and the use of the Wasserstein loss \cite{arjovsky2017wasserstein}, which improves the performance of GANs in several contexts \cite{weng2019gan, gulrajani2017improved}.
It would also be interesting to test MoGAN's effectiveness on cities of different sizes and shapes and regarding the generation of individual mobility trajectories, which represent the aggregated movements of single individuals among a city's locations \cite{moro2022, rinzivillo2014purpose, schneider2013unravelling}.
Finally, we plan to design a version of MoGAN capable of generating a network describing the mobility network of a weekday, a weekend day, or a specific day of the week.

An important aspect to investigate as future work is also to what extent MoGAN is geographically transferable \cite{luca2021survey}, i.e., it can be trained on a specific city and then used to generate mobility networks in a different city effectively. 
Geographic transferability can be crucial when there is a scarcity or even an absence of mobility data for a city. 

In this study, we use data from Chicago and New York City, which differ considerably by size, population, and shape, as well as by socio-demographic factors, POIs distribution, land use, etc. So, it does not make sense to transfer Chicago's MoGAN to New York City and vice versa. 
We leave experiments about the geographic transferability of MoGAN among cities to future works.

As MoGAN leverages the architecture of DCGAN, it only works with $64\times64$ matrices.
While representing geographic areas with less than $64\times64$ zones is not an issue (using, e.g., padding techniques \cite{goodfellow2016deep}), in its current version, MoGAN cannot work with areas split into more than $64\times64$ zones. 
Future extensions of MoGAN may consider using GAN architectures that deal with matrices larger than $64\times64$ \cite{wang2021generative}.
Adapting such models to deal with temporal and spatial aspects would allow us to design a new GAN for mobility flows to deal with larger geographic areas.

Another promising future direction is developing a GAN to generate a realistic mobility network for a specific condition (e.g., a rainy day or a day with some public events in the city). Having a so-called conditional GAN \cite{mirza2014conditional} may represent a unique opportunity for policymakers to generate realistic scenarios for specific circumstances. Finally, an exciting open challenge consists in interpreting which rules or well-known mobility laws (e.g., the gravity law) generative models are learning.

In the meantime, our study demonstrates the great potential of artificial intelligence to improve solutions to crucial problems in human mobility, such as the generation of realistic mobility networks. 
MoGAN can synthesize aggregated movements within a city into a realistic generator, which can be used for data augmentation, simulations, and what-if analysis.
Given the flexibility of the training phase, our model can be easily extended to synthesize specific types of mobility, such as aggregated movements during workdays, weekends, specific periods of the year, or in the presence of pandemic-driven mobility restrictions, events, and natural disasters.


\begin{backmatter}

\section*{Competing interests}
The authors declare that they have no competing interests.

\section*{Author's contributions}
GM: study conceptualization, data preprocessing and analysis, experiment running, code implementation, interpretation
of results, writing, plots and images. ML: study conceptualization, interpretation of results. AL: study conceptualization, interpretation of results, writing. BL: interpretation of results, writing, study direction. LP: study conceptualization, experiment design, interpretation of results, writing, study direction and management. 
All authors read and approved the final manuscript.

\section*{Availability of data and material}
The code to train/test MoGAN and reproduce our analyses, and the links to the datasets used in our experiments, can be found at \url{https://github.com/jonpappalord/GAN-flow}.

\section*{Funding}
Luca Pappalardo and Giovanni Mauro have been supported by EU projects: 1) SoBigData++ grant agreement \#871042 and 2) NextGenerationEU - National Recovery and Resilience Plan (Piano Nazionale di Ripresa e Resilienza, PNRR), project “SoBigData.it - Strengthening the Italian RI for Social Mining and Big Data Analytics”, prot. IR0000013, avviso n. 3264 on 28/12/2021.

\section*{Acknowledgments}
We thank Ramon Ferrer-i-Cancho, Matteo Böhm, Giuliano Cornacchia, and Vasiliki Voukelatou for the useful suggestions. We thank Daniele Fadda and Eleonora Cappuccio for the visualization suggestions. We thank CINI Lab for recognizing a price to the ideas that guided the development of MoGAN.
We also thank René Ferretti and Dante Milonga for the inspiration.

\bibliographystyle{bmc-mathphys} 
\bibliography{bmc_article}      


\begin{thebibliography}{65}
\ifx \bisbn   \undefined \def \bisbn  #1{ISBN #1}\fi
\ifx \binits  \undefined \def \binits#1{#1}\fi
\ifx \bauthor  \undefined \def \bauthor#1{#1}\fi
\ifx \batitle  \undefined \def \batitle#1{#1}\fi
\ifx \bjtitle  \undefined \def \bjtitle#1{#1}\fi
\ifx \bvolume  \undefined \def \bvolume#1{\textbf{#1}}\fi
\ifx \byear  \undefined \def \byear#1{#1}\fi
\ifx \bissue  \undefined \def \bissue#1{#1}\fi
\ifx \bfpage  \undefined \def \bfpage#1{#1}\fi
\ifx \blpage  \undefined \def \blpage #1{#1}\fi
\ifx \burl  \undefined \def \burl#1{\textsf{#1}}\fi
\ifx \doiurl  \undefined \def \doiurl#1{\textsf{#1}}\fi
\ifx \betal  \undefined \def \betal{\textit{et al.}}\fi
\ifx \binstitute  \undefined \def \binstitute#1{#1}\fi
\ifx \binstitutionaled  \undefined \def \binstitutionaled#1{#1}\fi
\ifx \bctitle  \undefined \def \bctitle#1{#1}\fi
\ifx \beditor  \undefined \def \beditor#1{#1}\fi
\ifx \bpublisher  \undefined \def \bpublisher#1{#1}\fi
\ifx \bbtitle  \undefined \def \bbtitle#1{#1}\fi
\ifx \bedition  \undefined \def \bedition#1{#1}\fi
\ifx \bseriesno  \undefined \def \bseriesno#1{#1}\fi
\ifx \blocation  \undefined \def \blocation#1{#1}\fi
\ifx \bsertitle  \undefined \def \bsertitle#1{#1}\fi
\ifx \bsnm \undefined \def \bsnm#1{#1}\fi
\ifx \bsuffix \undefined \def \bsuffix#1{#1}\fi
\ifx \bparticle \undefined \def \bparticle#1{#1}\fi
\ifx \barticle \undefined \def \barticle#1{#1}\fi
\ifx \bconfdate \undefined \def \bconfdate #1{#1}\fi
\ifx \botherref \undefined \def \botherref #1{#1}\fi
\ifx \url \undefined \def \url#1{\textsf{#1}}\fi
\ifx \bchapter \undefined \def \bchapter#1{#1}\fi
\ifx \bbook \undefined \def \bbook#1{#1}\fi
\ifx \bcomment \undefined \def \bcomment#1{#1}\fi
\ifx \oauthor \undefined \def \oauthor#1{#1}\fi
\ifx \citeauthoryear \undefined \def \citeauthoryear#1{#1}\fi
\ifx \endbibitem  \undefined \def \endbibitem {}\fi
\ifx \bconflocation  \undefined \def \bconflocation#1{#1}\fi
\ifx \arxivurl  \undefined \def \arxivurl#1{\textsf{#1}}\fi
\csname PreBibitemsHook\endcsname

\bibitem{batty2013new}
\begin{botherref}
\oauthor{\bsnm{Batty}, \binits{M.}}:
The new science of cities.
MIT press
(2013)
\end{botherref}
\endbibitem

\bibitem{andrienko2020sobigdata}
\begin{botherref}
\oauthor{\bsnm{Andrienko}, \binits{G.}},
\oauthor{\bsnm{Andrienko}, \binits{N.}},
\oauthor{\bsnm{Boldrini}, \binits{C.}},
\oauthor{\bsnm{Caldarelli}, \binits{G.}},
\oauthor{\bsnm{Cintia}, \binits{P.}},
\oauthor{\bsnm{Cresci}, \binits{S.}},
\oauthor{\bsnm{Facchini}, \binits{A.}},
\oauthor{\bsnm{Giannotti}, \binits{F.}},
\oauthor{\bsnm{Gionis}, \binits{A.}},
\oauthor{\bsnm{Guidotti}, \binits{R.}},
\oauthor{\bsnm{Mathioudakis}, \binits{M.}},
\oauthor{\bsnm{Muntean}, \binits{C.I.}},
\oauthor{\bsnm{Pappalardo}, \binits{L.}},
\oauthor{\bsnm{Pedreschi}, \binits{D.}},
\oauthor{\bsnm{Pournaras}, \binits{E.}},
\oauthor{\bsnm{Pratesi}, \binits{F.}},
\oauthor{\bsnm{Tesconi}, \binits{M.}},
\oauthor{\bsnm{Trasarti}, \binits{R.}}:
(so) big data and the transformation of the city.
International Journal of Data Science and Analytics
(2020)
\end{botherref}
\endbibitem

\bibitem{lucchini2021living}
\begin{barticle}
\bauthor{\bsnm{Lucchini}, \binits{L.}},
\bauthor{\bsnm{Centellegher}, \binits{S.}},
\bauthor{\bsnm{Pappalardo}, \binits{L.}},
\bauthor{\bsnm{Gallotti}, \binits{R.}},
\bauthor{\bsnm{Privitera}, \binits{F.}},
\bauthor{\bsnm{Lepri}, \binits{B.}},
\bauthor{\bsnm{{De Nadai}}, \binits{M.}}:
\batitle{{Living in a pandemic: changes in mobility routines, social activity
  and adherence to COVID-19 protective measures}}.
\bjtitle{Scientific Reports}
\bvolume{11}(\bissue{1}),
\bfpage{24452}
(\byear{2021}).
doi:\doiurl{10.1038/s41598-021-04139-1}
\end{barticle}
\endbibitem

\bibitem{pepe2020covid}
\begin{barticle}
\bauthor{\bsnm{Pepe}, \binits{E.}},
\bauthor{\bsnm{Bajardi}, \binits{P.}},
\bauthor{\bsnm{Gauvin}, \binits{L.}},
\bauthor{\bsnm{Privitera}, \binits{F.}},
\bauthor{\bsnm{Lake}, \binits{B.}},
\bauthor{\bsnm{Cattuto}, \binits{C.}},
\bauthor{\bsnm{Tizzoni}, \binits{M.}}:
\batitle{Covid-19 outbreak response, a dataset to assess mobility changes in
  italy following national lockdown}.
\bjtitle{Scientific data}
\bvolume{7}(\bissue{1}),
\bfpage{1}--\blpage{7}
(\byear{2020})
\end{barticle}
\endbibitem

\bibitem{lai2019measuring}
\begin{botherref}
\oauthor{\bsnm{Lai}, \binits{S.}},
\oauthor{\bsnm{Farnham}, \binits{A.}},
\oauthor{\bsnm{Ruktanonchai}, \binits{N.W.}},
\oauthor{\bsnm{Tatem}, \binits{A.J.}}:
Measuring mobility, disease connectivity and individual risk: a review of using
  mobile phone data and health for travel medicine.
Journal of travel medicine
\textbf{26}(3)
(2019)
\end{botherref}
\endbibitem

\bibitem{Ruktanonchai1465}
\begin{barticle}
\bauthor{\bsnm{Ruktanonchai}, \binits{N.W.}},
\bauthor{\bsnm{Floyd}, \binits{J.R.}},
\bauthor{\bsnm{Lai}, \binits{S.}},
\bauthor{\bsnm{Ruktanonchai}, \binits{C.W.}},
\bauthor{\bsnm{Sadilek}, \binits{A.}},
\bauthor{\bsnm{Rente-Lourenco}, \binits{P.}},
\bauthor{\bsnm{Ben}, \binits{X.}},
\bauthor{\bsnm{Carioli}, \binits{A.}},
\bauthor{\bsnm{Gwinn}, \binits{J.}},
\bauthor{\bsnm{Steele}, \binits{J.E.}},
\bauthor{\bsnm{Prosper}, \binits{O.}},
\bauthor{\bsnm{Schneider}, \binits{A.}},
\bauthor{\bsnm{Oplinger}, \binits{A.}},
\bauthor{\bsnm{Eastham}, \binits{P.}},
\bauthor{\bsnm{Tatem}, \binits{A.J.}}:
\batitle{Assessing the impact of coordinated covid-19 exit strategies across
  europe}.
\bjtitle{Science}
\bvolume{369}(\bissue{6510}),
\bfpage{1465}--\blpage{1470}
(\byear{2020})
\end{barticle}
\endbibitem

\bibitem{kraemer2020effect}
\begin{barticle}
\bauthor{\bsnm{Kraemer}, \binits{M.U.}},
\bauthor{\bsnm{Yang}, \binits{C.-H.}},
\bauthor{\bsnm{Gutierrez}, \binits{B.}},
\bauthor{\bsnm{Wu}, \binits{C.-H.}},
\bauthor{\bsnm{Klein}, \binits{B.}},
\bauthor{\bsnm{Pigott}, \binits{D.M.}},
\bauthor{\bsnm{Du~Plessis}, \binits{L.}},
\bauthor{\bsnm{Faria}, \binits{N.R.}},
\bauthor{\bsnm{Li}, \binits{R.}},
\bauthor{\bsnm{Hanage}, \binits{W.P.}}, \betal:
\batitle{The effect of human mobility and control measures on the covid-19
  epidemic in china}.
\bjtitle{Science}
\bvolume{368}(\bissue{6490}),
\bfpage{493}--\blpage{497}
(\byear{2020})
\end{barticle}
\endbibitem

\bibitem{oliver2020mobile}
\begin{botherref}
\oauthor{\bsnm{Oliver}, \binits{N.}},
\oauthor{\bsnm{Lepri}, \binits{B.}},
\oauthor{\bsnm{Sterly}, \binits{H.}},
\oauthor{\bsnm{Lambiotte}, \binits{R.}},
\oauthor{\bsnm{Deletaille}, \binits{S.}},
\oauthor{\bsnm{De~Nadai}, \binits{M.}},
\oauthor{\bsnm{Letouz{\'e}}, \binits{E.}},
\oauthor{\bsnm{Salah}, \binits{A.A.}},
\oauthor{\bsnm{Benjamins}, \binits{R.}},
\oauthor{\bsnm{Cattuto}, \binits{C.}}, et al.:
Mobile phone data for informing public health actions across the COVID-19
  pandemic life cycle
(2020)
\end{botherref}
\endbibitem

\bibitem{luca2021survey}
\begin{barticle}
\bauthor{\bsnm{Luca}, \binits{M.}},
\bauthor{\bsnm{Barlacchi}, \binits{G.}},
\bauthor{\bsnm{Lepri}, \binits{B.}},
\bauthor{\bsnm{Pappalardo}, \binits{L.}}:
\batitle{A survey on deep learning for human mobility}.
\bjtitle{ACM Computing Surveys (CSUR)}
\bvolume{55}(\bissue{1}),
\bfpage{1}--\blpage{44}
(\byear{2021})
\end{barticle}
\endbibitem

\bibitem{wang2019urban}
\begin{botherref}
\oauthor{\bsnm{Wang}, \binits{J.}},
\oauthor{\bsnm{Kong}, \binits{X.}},
\oauthor{\bsnm{Xia}, \binits{F.}},
\oauthor{\bsnm{Sun}, \binits{L.}}:
Urban human mobility: Data-driven modeling and prediction.
ACM SIGKDD Explorations Newsletter,
1--19
(2019)
\end{botherref}
\endbibitem

\bibitem{barbosa2018human}
\begin{barticle}
\bauthor{\bsnm{Barbosa}, \binits{H.}},
\bauthor{\bsnm{Barthelemy}, \binits{M.}},
\bauthor{\bsnm{Ghoshal}, \binits{G.}},
\bauthor{\bsnm{James}, \binits{C.R.}},
\bauthor{\bsnm{Lenormand}, \binits{M.}},
\bauthor{\bsnm{Louail}, \binits{T.}},
\bauthor{\bsnm{Menezes}, \binits{R.}},
\bauthor{\bsnm{Ramasco}, \binits{J.J.}},
\bauthor{\bsnm{Simini}, \binits{F.}},
\bauthor{\bsnm{Tomasini}, \binits{M.}}:
\batitle{{Human mobility: Models and applications}}.
\bjtitle{Physics Reports}
\bvolume{734},
\bfpage{1}--\blpage{74}
(\byear{2018}).
doi:\doiurl{10.1016/j.physrep.2018.01.001}
\end{barticle}
\endbibitem

\bibitem{le2015towards}
\begin{barticle}
\bauthor{\bsnm{Le~Blanc}, \binits{D.}}:
\batitle{Towards integration at last? the sustainable development goals as a
  network of targets}.
\bjtitle{Sustainable Development}
\bvolume{23}(\bissue{3}),
\bfpage{176}--\blpage{187}
(\byear{2015})
\end{barticle}
\endbibitem

\bibitem{kroll2019sustainable}
\begin{barticle}
\bauthor{\bsnm{Kroll}, \binits{C.}},
\bauthor{\bsnm{Warchold}, \binits{A.}},
\bauthor{\bsnm{Pradhan}, \binits{P.}}:
\batitle{Sustainable development goals (sdgs): Are we successful in turning
  trade-offs into synergies?}
\bjtitle{Palgrave Communications}
\bvolume{5}(\bissue{1}),
\bfpage{1}--\blpage{11}
(\byear{2019})
\end{barticle}
\endbibitem

\bibitem{assembly2015sustainable}
\begin{botherref}
\oauthor{\bsnm{{United Nations General Assembly}}}:
Transforming our world: the 2030 Agenda for Sustainable Development.
\url{https://sdgs.un.org/2030agenda}.
Accessed: 2021-02-23
(2015)
\end{botherref}
\endbibitem

\bibitem{bohm2022gross}
\begin{barticle}
\bauthor{\bsnm{B\"{o}hm}, \binits{M.}},
\bauthor{\bsnm{Nanni}, \binits{M.}},
\bauthor{\bsnm{Pappalardo}, \binits{L.}}:
\batitle{Gross polluters and vehicle emissions reduction}.
\bjtitle{Nature Sustainability}
(\byear{2022}).
doi:\doiurl{10.1038/s41893-022-00903-x}
\end{barticle}
\endbibitem

\bibitem{simini2021deep}
\begin{barticle}
\bauthor{\bsnm{Simini}, \binits{F.}},
\bauthor{\bsnm{Barlacchi}, \binits{G.}},
\bauthor{\bsnm{Luca}, \binits{M.}},
\bauthor{\bsnm{Pappalardo}, \binits{L.}}:
\batitle{A deep gravity model for mobility flows generation}.
\bjtitle{Nature Communications}
\bvolume{12}(\bissue{1}),
\bfpage{1}--\blpage{13}
(\byear{2021})
\end{barticle}
\endbibitem

\bibitem{masucci2013gravity}
\begin{barticle}
\bauthor{\bsnm{Masucci}, \binits{A.P.}},
\bauthor{\bsnm{Serras}, \binits{J.}},
\bauthor{\bsnm{Johansson}, \binits{A.}},
\bauthor{\bsnm{Batty}, \binits{M.}}:
\batitle{Gravity versus radiation models: On the importance of scale and
  heterogeneity in commuting flows}.
\bjtitle{Physical Review E}
\bvolume{88}(\bissue{2}),
\bfpage{022812}
(\byear{2013})
\end{barticle}
\endbibitem

\bibitem{carey1867principles}
\begin{botherref}
\oauthor{\bsnm{Carey}, \binits{H.C.}}:
Principles of social science.
JB Lippincott \& Company
(1867)
\end{botherref}
\endbibitem

\bibitem{zipf1946p}
\begin{barticle}
\bauthor{\bsnm{Zipf}, \binits{G.K.}}:
\batitle{The p 1 p 2/d hypothesis: on the intercity movement of persons}.
\bjtitle{American sociological review}
\bvolume{11}(\bissue{6}),
\bfpage{677}--\blpage{686}
(\byear{1946})
\end{barticle}
\endbibitem

\bibitem{lenormand2016systematic}
\begin{barticle}
\bauthor{\bsnm{Lenormand}, \binits{M.}},
\bauthor{\bsnm{Bassolas}, \binits{A.}},
\bauthor{\bsnm{Ramasco}, \binits{J.J.}}:
\batitle{Systematic comparison of trip distribution laws and models}.
\bjtitle{Journal of Transport Geography}
\bvolume{51},
\bfpage{158}--\blpage{169}
(\byear{2016})
\end{barticle}
\endbibitem

\bibitem{erlander1990gravity}
\begin{botherref}
\oauthor{\bsnm{Erlander}, \binits{S.}},
\oauthor{\bsnm{Stewart}, \binits{N.F.}}:
The gravity model in transportation analysis: theory and extensions.
Vsp
(1990)
\end{botherref}
\endbibitem

\bibitem{luca2022modeling}
\begin{barticle}
\bauthor{\bsnm{Luca}, \binits{M.}},
\bauthor{\bsnm{Lepri}, \binits{B.}},
\bauthor{\bsnm{Frias-Martinez}, \binits{E.}},
\bauthor{\bsnm{Lutu}, \binits{A.}}:
\batitle{Modeling international mobility using roaming cell phone traces during
  covid-19 pandemic}.
\bjtitle{EPJ Data Science}
\bvolume{11}(\bissue{1}),
\bfpage{22}
(\byear{2022})
\end{barticle}
\endbibitem

\bibitem{simini2012universal}
\begin{barticle}
\bauthor{\bsnm{Simini}, \binits{F.}},
\bauthor{\bsnm{Gonz{\'a}lez}, \binits{M.C.}},
\bauthor{\bsnm{Maritan}, \binits{A.}},
\bauthor{\bsnm{Barab{\'a}si}, \binits{A.-L.}}:
\batitle{A universal model for mobility and migration patterns}.
\bjtitle{Nature}
\bvolume{484}(\bissue{7392}),
\bfpage{96}--\blpage{100}
(\byear{2012})
\end{barticle}
\endbibitem

\bibitem{prieto2018gravity}
\begin{barticle}
\bauthor{\bsnm{{Prieto Curiel}}, \binits{R.}},
\bauthor{\bsnm{Pappalardo}, \binits{L.}},
\bauthor{\bsnm{Gabrielli}, \binits{L.}},
\bauthor{\bsnm{Bishop}, \binits{S.R.}}:
\batitle{{Gravity and scaling laws of city to city migration}}.
\bjtitle{PLOS ONE}
\bvolume{13}(\bissue{7}),
\bfpage{1}--\blpage{19}
(\byear{2018}).
doi:\doiurl{10.1371/journal.pone.0199892}
\end{barticle}
\endbibitem

\bibitem{yan2017universal}
\begin{barticle}
\bauthor{\bsnm{Yan}, \binits{X.-Y.}},
\bauthor{\bsnm{Wang}, \binits{W.-X.}},
\bauthor{\bsnm{Gao}, \binits{Z.-Y.}},
\bauthor{\bsnm{Lai}, \binits{Y.-C.}}:
\batitle{{Universal model of individual and population mobility on diverse
  spatial scales}}.
\bjtitle{Nature Communications}
\bvolume{8}(\bissue{1}),
\bfpage{1639}
(\byear{2017}).
doi:\doiurl{10.1038/s41467-017-01892-8}
\end{barticle}
\endbibitem

\bibitem{goodfellow2014generative}
\begin{bchapter}
\bauthor{\bsnm{Goodfellow}, \binits{I.J.}},
\bauthor{\bsnm{Pouget-Abadie}, \binits{J.}},
\bauthor{\bsnm{Mirza}, \binits{M.}},
\bauthor{\bsnm{Xu}, \binits{B.}},
\bauthor{\bsnm{Warde-Farley}, \binits{D.}},
\bauthor{\bsnm{Ozair}, \binits{S.}},
\bauthor{\bsnm{Courville}, \binits{A.}},
\bauthor{\bsnm{Bengio}, \binits{Y.}}:
\bctitle{Generative adversarial nets}.
In: \bbtitle{Proceedings of the 27th International Conference on Neural
  Information Processing Systems},
pp. \bfpage{2672}--\blpage{2680}
(\byear{2014})
\end{bchapter}
\endbibitem

\bibitem{creswell2018generative}
\begin{barticle}
\bauthor{\bsnm{Creswell}, \binits{A.}},
\bauthor{\bsnm{White}, \binits{T.}},
\bauthor{\bsnm{Dumoulin}, \binits{V.}},
\bauthor{\bsnm{Arulkumaran}, \binits{K.}},
\bauthor{\bsnm{Sengupta}, \binits{B.}},
\bauthor{\bsnm{Bharath}, \binits{A.A.}}:
\batitle{Generative adversarial networks: An overview}.
\bjtitle{IEEE Signal Processing Magazine}
\bvolume{35}(\bissue{1}),
\bfpage{53}--\blpage{65}
(\byear{2018})
\end{barticle}
\endbibitem

\bibitem{goodfellow2016nips}
\begin{botherref}
\oauthor{\bsnm{Goodfellow}, \binits{I.}}:
Nips 2016 tutorial: Generative adversarial networks.
arXiv preprint arXiv:1701.00160
(2016)
\end{botherref}
\endbibitem

\bibitem{radford2016unsupervised}
\begin{botherref}
\oauthor{\bsnm{Radford}, \binits{A.}},
\oauthor{\bsnm{Metz}, \binits{L.}},
\oauthor{\bsnm{Chintala}, \binits{S.}}:
Unsupervised Representation Learning with Deep Convolutional Generative
  Adversarial Networks
(2016).
\arxivurl{1511.06434}
\end{botherref}
\endbibitem

\bibitem{liu2018trajgans}
\begin{bchapter}
\bauthor{\bsnm{Liu}, \binits{X.}},
\bauthor{\bsnm{Chen}, \binits{H.}},
\bauthor{\bsnm{Andris}, \binits{C.}}:
\bctitle{trajgans: Using generative adversarial networks for geo-privacy
  protection of trajectory data (vision paper)}.
In: \bbtitle{Location Privacy and Security Workshop},
pp. \bfpage{1}--\blpage{7}
(\byear{2018})
\end{bchapter}
\endbibitem

\bibitem{yin2018gans}
\begin{botherref}
\oauthor{\bsnm{Yin}, \binits{D.}},
\oauthor{\bsnm{Yang}, \binits{Q.}}:
Gans based density distribution privacy-preservation on mobility data.
Security and Communication Networks
\textbf{2018}
(2018)
\end{botherref}
\endbibitem

\bibitem{kulkarni2018generative}
\begin{botherref}
\oauthor{\bsnm{Kulkarni}, \binits{V.}},
\oauthor{\bsnm{Tagasovska}, \binits{N.}},
\oauthor{\bsnm{Vatter}, \binits{T.}},
\oauthor{\bsnm{Garbinato}, \binits{B.}}:
Generative models for simulating mobility trajectories.
arXiv preprint arXiv:1811.12801
(2018)
\end{botherref}
\endbibitem

\bibitem{ouyang2018non}
\begin{bchapter}
\bauthor{\bsnm{Ouyang}, \binits{K.}},
\bauthor{\bsnm{Shokri}, \binits{R.}},
\bauthor{\bsnm{Rosenblum}, \binits{D.S.}},
\bauthor{\bsnm{Yang}, \binits{W.}}:
\bctitle{A non-parametric generative model for human trajectories.}
In: \bbtitle{IJCAI},
pp. \bfpage{3812}--\blpage{3817}
(\byear{2018})
\end{bchapter}
\endbibitem

\bibitem{huang2019variational}
\begin{bchapter}
\bauthor{\bsnm{{Huang}}, \binits{D.}},
\bauthor{\bsnm{{Song}}, \binits{X.}},
\bauthor{\bsnm{{Fan}}, \binits{Z.}},
\bauthor{\bsnm{{Jiang}}, \binits{R.}},
\bauthor{\bsnm{{Shibasaki}}, \binits{R.}},
\bauthor{\bsnm{{Zhang}}, \binits{Y.}},
\bauthor{\bsnm{{Wang}}, \binits{H.}},
\bauthor{\bsnm{{Kato}}, \binits{Y.}}:
\bctitle{A variational autoencoder based generative model of urban human
  mobility}.
In: \bbtitle{2019 IEEE Conference on Multimedia Information Processing and
  Retrieval (MIPR)},
pp. \bfpage{425}--\blpage{430}
(\byear{2019})
\end{bchapter}
\endbibitem

\bibitem{feng2020learning}
\begin{bchapter}
\bauthor{\bsnm{Feng}, \binits{J.}},
\bauthor{\bsnm{Yang}, \binits{Z.}},
\bauthor{\bsnm{Xu}, \binits{F.}},
\bauthor{\bsnm{Yu}, \binits{H.}},
\bauthor{\bsnm{Wang}, \binits{M.}},
\bauthor{\bsnm{Li}, \binits{Y.}}:
\bctitle{Learning to simulate human mobility}.
In: \bbtitle{Proceedings of the 26th ACM SIGKDD International Conference on
  Knowledge Discovery \& Data Mining}.
\bsertitle{KDD '20},
pp. \bfpage{3426}--\blpage{3433}.
\bpublisher{Association for Computing Machinery},
\blocation{New York, NY, USA}
(\byear{2020}).
doi:\doiurl{10.1145/3394486.3412862}.
\burl{https://doi.org/10.1145/3394486.3412862}
\end{bchapter}
\endbibitem

\bibitem{moreira2013predicting}
\begin{barticle}
\bauthor{\bsnm{Moreira-Matias}, \binits{L.}},
\bauthor{\bsnm{Gama}, \binits{J.}},
\bauthor{\bsnm{Ferreira}, \binits{M.}},
\bauthor{\bsnm{Mendes-Moreira}, \binits{J.}},
\bauthor{\bsnm{Damas}, \binits{L.}}:
\batitle{Predicting taxi--passenger demand using streaming data}.
\bjtitle{IEEE Transactions on Intelligent Transportation Systems}
\bvolume{14}(\bissue{3}),
\bfpage{1393}--\blpage{1402}
(\byear{2013})
\end{barticle}
\endbibitem

\bibitem{cornacchia2021modelling}
\begin{botherref}
\oauthor{\bsnm{Cornacchia}, \binits{G.}},
\oauthor{\bsnm{Pappalardo}, \binits{L.}}:
A mechanistic data-driven approach to synthesize human mobility considering the
  spatial, temporal, and social dimensions together.
ISPRS International Journal of Geo-Information
\textbf{10}(9)
(2021).
doi:\doiurl{10.3390/ijgi10090599}
\end{botherref}
\endbibitem

\bibitem{pappalardo2017data}
\begin{barticle}
\bauthor{\bsnm{Pappalardo}, \binits{L.}},
\bauthor{\bsnm{Simini}, \binits{F.}}:
\batitle{Data-driven generation of spatio-temporal routines in human mobility}.
\bjtitle{Data Mining and Knowledge Discovery}
\bvolume{32}(\bissue{3}),
\bfpage{787}--\blpage{829}
(\byear{2018})
\end{barticle}
\endbibitem

\bibitem{jiang2016timegeo}
\begin{barticle}
\bauthor{\bsnm{Jiang}, \binits{S.}},
\bauthor{\bsnm{Yang}, \binits{Y.}},
\bauthor{\bsnm{Gupta}, \binits{S.}},
\bauthor{\bsnm{Veneziano}, \binits{D.}},
\bauthor{\bsnm{Athavale}, \binits{S.}},
\bauthor{\bsnm{Gonzalez}, \binits{M.C.}}:
\batitle{The timegeo modeling framework for urban mobility without travel
  surveys}.
\bjtitle{Proceedings of the National Academy of Sciences}
\bvolume{113},
\bfpage{201524261}
(\byear{2016})
\end{barticle}
\endbibitem

\bibitem{pappalardo2019scikitmobility}
\begin{barticle}
\bauthor{\bsnm{Pappalardo}, \binits{L.}},
\bauthor{\bsnm{Simini}, \binits{F.}},
\bauthor{\bsnm{Barlacchi}, \binits{G.}},
\bauthor{\bsnm{Pellungrini}, \binits{R.}}:
\batitle{scikit-mobility: A python library for the analysis, generation, and
  risk assessment of mobility data}.
\bjtitle{Journal of Statistical Software}
\bvolume{103}(\bissue{4}),
\bfpage{1}--\blpage{38}
(\byear{2022}).
doi:\doiurl{10.18637/jss.v103.i04}
\end{barticle}
\endbibitem

\bibitem{citi}
\begin{botherref}
\oauthor{\bsnm{citibikenyc.com}}:
CitiBike System Data
(2013--).
\url{https://www.citibikenyc.com/system-data}
\end{botherref}
\endbibitem

\bibitem{divvy}
\begin{botherref}
\oauthor{\bsnm{divvybikes.com}}:
Divvy System Data
(2016--).
\url{https://www.divvybikes.com/system-data}
\end{botherref}
\endbibitem

\bibitem{taxi_nyc}
\begin{botherref}
\oauthor{\bsnm{nyc.gov}}:
TLC Trip Record Data
(2009--).
\url{https://www1.nyc.gov/site/tlc/about/tlc-trip-record-data.page}
\end{botherref}
\endbibitem

\bibitem{taxi_chi}
\begin{botherref}
\oauthor{\bsnm{chicago.gov}}:
TLC Trip Record Data
(2013--).
\url{https://data.cityofchicago.org/Transportation/Taxi-Trips/wrvz-psew/data}
\end{botherref}
\endbibitem

\bibitem{tantardini2019comparing}
\begin{barticle}
\bauthor{\bsnm{Tantardini}, \binits{M.}},
\bauthor{\bsnm{Ieva}, \binits{F.}},
\bauthor{\bsnm{Tajoli}, \binits{L.}},
\bauthor{\bsnm{Piccardi}, \binits{C.}}:
\batitle{Comparing methods for comparing networks}.
\bjtitle{Scientific reports}
\bvolume{9}(\bissue{1}),
\bfpage{1}--\blpage{19}
(\byear{2019})
\end{barticle}
\endbibitem

\bibitem{LENORMAND2016158}
\begin{barticle}
\bauthor{\bsnm{Lenormand}, \binits{M.}},
\bauthor{\bsnm{Bassolas}, \binits{A.}},
\bauthor{\bsnm{Ramasco}, \binits{J.J.}}:
\batitle{{Systematic comparison of trip distribution laws and models}}.
\bjtitle{Journal of Transport Geography}
\bvolume{51},
\bfpage{158}--\blpage{169}
(\byear{2016}).
doi:\doiurl{10.1016/j.jtrangeo.2015.12.008}
\end{barticle}
\endbibitem

\bibitem{liu2018cut}
\begin{barticle}
\bauthor{\bsnm{Liu}, \binits{Q.}},
\bauthor{\bsnm{Dong}, \binits{Z.}},
\bauthor{\bsnm{Wang}, \binits{E.}}:
\batitle{Cut based method for comparing complex networks}.
\bjtitle{Scientific reports}
\bvolume{8}(\bissue{1}),
\bfpage{1}--\blpage{11}
(\byear{2018})
\end{barticle}
\endbibitem

\bibitem{alon2004approximating}
\begin{bchapter}
\bauthor{\bsnm{Alon}, \binits{N.}},
\bauthor{\bsnm{Naor}, \binits{A.}}:
\bctitle{Approximating the cut-norm via grothendieck's inequality}.
In: \bbtitle{Proceedings of the Thirty-sixth Annual ACM Symposium on Theory of
  Computing},
pp. \bfpage{72}--\blpage{80}
(\byear{2004})
\end{bchapter}
\endbibitem

\bibitem{o2008optimal}
\begin{bchapter}
\bauthor{\bsnm{O'Donnell}, \binits{R.}},
\bauthor{\bsnm{Wu}, \binits{Y.}}:
\bctitle{An optimal sdp algorithm for max-cut, and equally optimal long code
  tests}.
In: \bbtitle{Proceedings of the Fortieth Annual ACM Symposium on Theory of
  Computing},
pp. \bfpage{335}--\blpage{344}
(\byear{2008})
\end{bchapter}
\endbibitem

\bibitem{cutnorm}
\begin{botherref}
\oauthor{\bsnm{Chiu}, \binits{P.-K.}}:
cutnorm package
(2018--).
\url{https://pypi.org/project/cutnorm/}
\end{botherref}
\endbibitem

\bibitem{fuglede2004jensen}
\begin{bchapter}
\bauthor{\bsnm{Fuglede}, \binits{B.}},
\bauthor{\bsnm{Topsoe}, \binits{F.}}:
\bctitle{Jensen-shannon divergence and hilbert space embedding}.
In: \bbtitle{International Symposium onInformation Theory, 2004. ISIT 2004.
  Proceedings.},
p. \bfpage{31}
(\byear{2004}).
\bcomment{IEEE}
\end{bchapter}
\endbibitem

\bibitem{kullback1997information}
\begin{botherref}
\oauthor{\bsnm{Kullback}, \binits{S.}}:
Information theory and statistics.
Courier Corporation
(1997)
\end{botherref}
\endbibitem

\bibitem{van2014renyi}
\begin{barticle}
\bauthor{\bsnm{Van~Erven}, \binits{T.}},
\bauthor{\bsnm{Harremos}, \binits{P.}}:
\batitle{R{\'e}nyi divergence and kullback-leibler divergence}.
\bjtitle{IEEE Transactions on Information Theory}
\bvolume{60}(\bissue{7}),
\bfpage{3797}--\blpage{3820}
(\byear{2014})
\end{barticle}
\endbibitem

\bibitem{vishwanathan2010}
\begin{barticle}
\bauthor{\bsnm{Vishwanathan}, \binits{S.V.N.}},
\bauthor{\bsnm{Schraudolph}, \binits{N.N.}},
\bauthor{\bsnm{Kondor}, \binits{R.}},
\bauthor{\bsnm{Borgwardt}, \binits{K.M.}}:
\batitle{Graph kernels}.
\bjtitle{Journal of Machine Learning Research}
\bvolume{11},
\bfpage{1201}--\blpage{1242}
(\byear{2010})
\end{barticle}
\endbibitem

\bibitem{nikolentzos2021}
\begin{barticle}
\bauthor{\bsnm{Nikolentzos}, \binits{G.}},
\bauthor{\bsnm{Siglidis}, \binits{G.}},
\bauthor{\bsnm{Vazirgiannis}, \binits{M.}}:
\batitle{Graph kernels: A survey}.
\bjtitle{Journal of Artificial Intelligence Research}
\bvolume{72},
\bfpage{943}--\blpage{1027}
(\byear{2021})
\end{barticle}
\endbibitem

\bibitem{scarselli2008graph}
\begin{barticle}
\bauthor{\bsnm{Scarselli}, \binits{F.}},
\bauthor{\bsnm{Gori}, \binits{M.}},
\bauthor{\bsnm{Tsoi}, \binits{A.C.}},
\bauthor{\bsnm{Hagenbuchner}, \binits{M.}},
\bauthor{\bsnm{Monfardini}, \binits{G.}}:
\batitle{The graph neural network model}.
\bjtitle{IEEE transactions on neural networks}
\bvolume{20}(\bissue{1}),
\bfpage{61}--\blpage{80}
(\byear{2008})
\end{barticle}
\endbibitem

\bibitem{arjovsky2017wasserstein}
\begin{bchapter}
\bauthor{\bsnm{Arjovsky}, \binits{M.}},
\bauthor{\bsnm{Chintala}, \binits{S.}},
\bauthor{\bsnm{Bottou}, \binits{L.}}:
\bctitle{Wasserstein generative adversarial networks}.
In: \bbtitle{International Conference on Machine Learning},
pp. \bfpage{214}--\blpage{223}
(\byear{2017}).
\bcomment{PMLR}
\end{bchapter}
\endbibitem

\bibitem{weng2019gan}
\begin{botherref}
\oauthor{\bsnm{Weng}, \binits{L.}}:
From gan to wgan.
arXiv preprint arXiv:1904.08994
(2019)
\end{botherref}
\endbibitem

\bibitem{gulrajani2017improved}
\begin{bchapter}
\bauthor{\bsnm{Gulrajani}, \binits{I.}},
\bauthor{\bsnm{Ahmed}, \binits{F.}},
\bauthor{\bsnm{Arjovsky}, \binits{M.}},
\bauthor{\bsnm{Dumoulin}, \binits{V.}},
\bauthor{\bsnm{Courville}, \binits{A.}}:
\bctitle{Improved training of wasserstein gans}.
In: \bbtitle{Proceedings of the 31st International Conference on Neural
  Information Processing Systems},
pp. \bfpage{5769}--\blpage{5779}
(\byear{2017})
\end{bchapter}
\endbibitem

\bibitem{moro2022}
\begin{botherref}
\oauthor{\bsnm{Berke}, \binits{A.}},
\oauthor{\bsnm{Doorley}, \binits{R.}},
\oauthor{\bsnm{Larson}, \binits{K.}},
\oauthor{\bsnm{Moro}, \binits{E.}}:
Generating synthetic mobility data for a realistic population with rnns to
  improve utility and privacy.
CoRR
\textbf{abs/2201.01139}
(2022).
\arxivurl{2201.01139}
\end{botherref}
\endbibitem

\bibitem{rinzivillo2014purpose}
\begin{bchapter}
\bauthor{\bsnm{{Rinzivillo}}, \binits{S.}},
\bauthor{\bsnm{{Gabrielli}}, \binits{L.}},
\bauthor{\bsnm{{Nanni}}, \binits{M.}},
\bauthor{\bsnm{{Pappalardo}}, \binits{L.}},
\bauthor{\bsnm{{Pedreschi}}, \binits{D.}},
\bauthor{\bsnm{{Giannotti}}, \binits{F.}}:
\bctitle{The purpose of motion: Learning activities from individual mobility
  networks}.
In: \bbtitle{2014 International Conference on Data Science and Advanced
  Analytics (DSAA)},
pp. \bfpage{312}--\blpage{318}
(\byear{2014})
\end{bchapter}
\endbibitem

\bibitem{schneider2013unravelling}
\begin{barticle}
\bauthor{\bsnm{Schneider}, \binits{C.M.}},
\bauthor{\bsnm{Belik}, \binits{V.}},
\bauthor{\bsnm{Couronné}, \binits{T.}},
\bauthor{\bsnm{Smoreda}, \binits{Z.}},
\bauthor{\bsnm{González}, \binits{M.C.}}:
\batitle{Unravelling daily human mobility motifs}.
\bjtitle{Journal of The Royal Society Interface}
\bvolume{10}(\bissue{84}),
\bfpage{20130246}
(\byear{2013})
\end{barticle}
\endbibitem

\bibitem{goodfellow2016deep}
\begin{bbook}
\bauthor{\bsnm{Goodfellow}, \binits{I.}},
\bauthor{\bsnm{Bengio}, \binits{Y.}},
\bauthor{\bsnm{Courville}, \binits{A.}}:
\bbtitle{Deep Learning}.
\bpublisher{MIT press}, \blocation{???}
(\byear{2016})
\end{bbook}
\endbibitem

\bibitem{wang2021generative}
\begin{barticle}
\bauthor{\bsnm{Wang}, \binits{Z.}},
\bauthor{\bsnm{She}, \binits{Q.}},
\bauthor{\bsnm{Ward}, \binits{T.E.}}:
\batitle{Generative adversarial networks in computer vision: A survey and
  taxonomy}.
\bjtitle{ACM Computing Surveys (CSUR)}
\bvolume{54}(\bissue{2}),
\bfpage{1}--\blpage{38}
(\byear{2021})
\end{barticle}
\endbibitem

\bibitem{mirza2014conditional}
\begin{botherref}
\oauthor{\bsnm{Mirza}, \binits{M.}},
\oauthor{\bsnm{Osindero}, \binits{S.}}:
Conditional generative adversarial nets.
arXiv preprint arXiv:1411.1784
(2014)
\end{botherref}
\endbibitem

\end{thebibliography}

\newcommand{\BMCxmlcomment}[1]{}

\BMCxmlcomment{

<refgrp>

<bibl id="B1">
  <title><p>The new science of cities</p></title>
  <aug>
    <au><snm>Batty</snm><fnm>M</fnm></au>
  </aug>
  <publisher>MIT press</publisher>
  <pubdate>2013</pubdate>
</bibl>

<bibl id="B2">
  <title><p>(So) Big Data and the transformation of the city</p></title>
  <aug>
    <au><snm>Andrienko</snm><fnm>G</fnm></au>
    <au><snm>Andrienko</snm><fnm>N</fnm></au>
    <au><snm>Boldrini</snm><fnm>C</fnm></au>
    <au><snm>Caldarelli</snm><fnm>G</fnm></au>
    <au><snm>Cintia</snm><fnm>P</fnm></au>
    <au><snm>Cresci</snm><fnm>S</fnm></au>
    <au><snm>Facchini</snm><fnm>A</fnm></au>
    <au><snm>Giannotti</snm><fnm>F</fnm></au>
    <au><snm>Gionis</snm><fnm>A</fnm></au>
    <au><snm>Guidotti</snm><fnm>R</fnm></au>
    <au><snm>Mathioudakis</snm><fnm>M</fnm></au>
    <au><snm>Muntean</snm><fnm>CI</fnm></au>
    <au><snm>Pappalardo</snm><fnm>L</fnm></au>
    <au><snm>Pedreschi</snm><fnm>D</fnm></au>
    <au><snm>Pournaras</snm><fnm>E</fnm></au>
    <au><snm>Pratesi</snm><fnm>F</fnm></au>
    <au><snm>Tesconi</snm><fnm>M</fnm></au>
    <au><snm>Trasarti</snm><fnm>R</fnm></au>
  </aug>
  <source>International Journal of Data Science and Analytics</source>
  <pubdate>2020</pubdate>
</bibl>

<bibl id="B3">
  <title><p>{Living in a pandemic: changes in mobility routines, social
  activity and adherence to COVID-19 protective measures}</p></title>
  <aug>
    <au><snm>Lucchini</snm><fnm>L</fnm></au>
    <au><snm>Centellegher</snm><fnm>S</fnm></au>
    <au><snm>Pappalardo</snm><fnm>L</fnm></au>
    <au><snm>Gallotti</snm><fnm>R</fnm></au>
    <au><snm>Privitera</snm><fnm>F</fnm></au>
    <au><snm>Lepri</snm><fnm>B</fnm></au>
    <au><snm>{De Nadai}</snm><fnm>M</fnm></au>
  </aug>
  <source>Scientific Reports</source>
  <pubdate>2021</pubdate>
  <volume>11</volume>
  <issue>1</issue>
  <fpage>24452</fpage>
  <url>https://doi.org/10.1038/s41598-021-04139-1</url>
</bibl>

<bibl id="B4">
  <title><p>COVID-19 outbreak response, a dataset to assess mobility changes in
  Italy following national lockdown</p></title>
  <aug>
    <au><snm>Pepe</snm><fnm>E</fnm></au>
    <au><snm>Bajardi</snm><fnm>P</fnm></au>
    <au><snm>Gauvin</snm><fnm>L</fnm></au>
    <au><snm>Privitera</snm><fnm>F</fnm></au>
    <au><snm>Lake</snm><fnm>B</fnm></au>
    <au><snm>Cattuto</snm><fnm>C</fnm></au>
    <au><snm>Tizzoni</snm><fnm>M</fnm></au>
  </aug>
  <source>Scientific data</source>
  <publisher>Nature Publishing Group</publisher>
  <pubdate>2020</pubdate>
  <volume>7</volume>
  <issue>1</issue>
  <fpage>1</fpage>
  <lpage>-7</lpage>
</bibl>

<bibl id="B5">
  <title><p>Measuring mobility, disease connectivity and individual risk: a
  review of using mobile phone data and Health for travel medicine</p></title>
  <aug>
    <au><snm>Lai</snm><fnm>S</fnm></au>
    <au><snm>Farnham</snm><fnm>A</fnm></au>
    <au><snm>Ruktanonchai</snm><fnm>NW</fnm></au>
    <au><snm>Tatem</snm><fnm>AJ</fnm></au>
  </aug>
  <source>Journal of travel medicine</source>
  <pubdate>2019</pubdate>
  <volume>26</volume>
  <issue>3</issue>
</bibl>

<bibl id="B6">
  <title><p>Assessing the impact of coordinated COVID-19 exit strategies across
  Europe</p></title>
  <aug>
    <au><snm>Ruktanonchai</snm><fnm>N. W.</fnm></au>
    <au><snm>Floyd</snm><fnm>J. R.</fnm></au>
    <au><snm>Lai</snm><fnm>S.</fnm></au>
    <au><snm>Ruktanonchai</snm><fnm>C. W.</fnm></au>
    <au><snm>Sadilek</snm><fnm>A.</fnm></au>
    <au><snm>Rente Lourenco</snm><fnm>P.</fnm></au>
    <au><snm>Ben</snm><fnm>X.</fnm></au>
    <au><snm>Carioli</snm><fnm>A.</fnm></au>
    <au><snm>Gwinn</snm><fnm>J.</fnm></au>
    <au><snm>Steele</snm><fnm>J. E.</fnm></au>
    <au><snm>Prosper</snm><fnm>O.</fnm></au>
    <au><snm>Schneider</snm><fnm>A.</fnm></au>
    <au><snm>Oplinger</snm><fnm>A.</fnm></au>
    <au><snm>Eastham</snm><fnm>P.</fnm></au>
    <au><snm>Tatem</snm><fnm>A. J.</fnm></au>
  </aug>
  <source>Science</source>
  <publisher>American Association for the Advancement of Science</publisher>
  <pubdate>2020</pubdate>
  <volume>369</volume>
  <issue>6510</issue>
  <fpage>1465</fpage>
  <lpage>-1470</lpage>
</bibl>

<bibl id="B7">
  <title><p>The effect of human mobility and control measures on the COVID-19
  epidemic in China</p></title>
  <aug>
    <au><snm>Kraemer</snm><fnm>MU</fnm></au>
    <au><snm>Yang</snm><fnm>CH</fnm></au>
    <au><snm>Gutierrez</snm><fnm>B</fnm></au>
    <au><snm>Wu</snm><fnm>CH</fnm></au>
    <au><snm>Klein</snm><fnm>B</fnm></au>
    <au><snm>Pigott</snm><fnm>DM</fnm></au>
    <au><snm>Du Plessis</snm><fnm>L</fnm></au>
    <au><snm>Faria</snm><fnm>NR</fnm></au>
    <au><snm>Li</snm><fnm>R</fnm></au>
    <au><snm>Hanage</snm><fnm>WP</fnm></au>
    <au><cnm>others</cnm></au>
  </aug>
  <source>Science</source>
  <publisher>American Association for the Advancement of Science</publisher>
  <pubdate>2020</pubdate>
  <volume>368</volume>
  <issue>6490</issue>
  <fpage>493</fpage>
  <lpage>-497</lpage>
</bibl>

<bibl id="B8">
  <title><p>Mobile phone data for informing public health actions across the
  COVID-19 pandemic life cycle</p></title>
  <aug>
    <au><snm>Oliver</snm><fnm>N</fnm></au>
    <au><snm>Lepri</snm><fnm>B</fnm></au>
    <au><snm>Sterly</snm><fnm>H</fnm></au>
    <au><snm>Lambiotte</snm><fnm>R</fnm></au>
    <au><snm>Deletaille</snm><fnm>S</fnm></au>
    <au><snm>De Nadai</snm><fnm>M</fnm></au>
    <au><snm>Letouz{\'e}</snm><fnm>E</fnm></au>
    <au><snm>Salah</snm><fnm>AA</fnm></au>
    <au><snm>Benjamins</snm><fnm>R</fnm></au>
    <au><snm>Cattuto</snm><fnm>C</fnm></au>
    <au><cnm>others</cnm></au>
  </aug>
  <source>Science Advances</source>
  <pubdate>2020</pubdate>
  <issue>23</issue>
</bibl>

<bibl id="B9">
  <title><p>A Survey on Deep Learning for Human Mobility</p></title>
  <aug>
    <au><snm>Luca</snm><fnm>M</fnm></au>
    <au><snm>Barlacchi</snm><fnm>G</fnm></au>
    <au><snm>Lepri</snm><fnm>B</fnm></au>
    <au><snm>Pappalardo</snm><fnm>L</fnm></au>
  </aug>
  <source>ACM Computing Surveys (CSUR)</source>
  <publisher>ACM New York, NY</publisher>
  <pubdate>2021</pubdate>
  <volume>55</volume>
  <issue>1</issue>
  <fpage>1</fpage>
  <lpage>-44</lpage>
</bibl>

<bibl id="B10">
  <title><p>Urban human mobility: Data-driven modeling and
  prediction</p></title>
  <aug>
    <au><snm>Wang</snm><fnm>J</fnm></au>
    <au><snm>Kong</snm><fnm>X</fnm></au>
    <au><snm>Xia</snm><fnm>F</fnm></au>
    <au><snm>Sun</snm><fnm>L</fnm></au>
  </aug>
  <source>ACM SIGKDD Explorations Newsletter</source>
  <pubdate>2019</pubdate>
  <fpage>1</fpage>
  <lpage>-19</lpage>
</bibl>

<bibl id="B11">
  <title><p>{Human mobility: Models and applications}</p></title>
  <aug>
    <au><snm>Barbosa</snm><fnm>H</fnm></au>
    <au><snm>Barthelemy</snm><fnm>M</fnm></au>
    <au><snm>Ghoshal</snm><fnm>G</fnm></au>
    <au><snm>James</snm><fnm>CR</fnm></au>
    <au><snm>Lenormand</snm><fnm>M</fnm></au>
    <au><snm>Louail</snm><fnm>T</fnm></au>
    <au><snm>Menezes</snm><fnm>R</fnm></au>
    <au><snm>Ramasco</snm><fnm>JJ</fnm></au>
    <au><snm>Simini</snm><fnm>F</fnm></au>
    <au><snm>Tomasini</snm><fnm>M</fnm></au>
  </aug>
  <source>Physics Reports</source>
  <pubdate>2018</pubdate>
  <volume>734</volume>
  <fpage>1</fpage>
  <lpage>-74</lpage>
  <url>https://www.sciencedirect.com/science/article/pii/S037015731830022X</url>
</bibl>

<bibl id="B12">
  <title><p>Towards integration at last? The sustainable development goals as a
  network of targets</p></title>
  <aug>
    <au><snm>Le Blanc</snm><fnm>D</fnm></au>
  </aug>
  <source>Sustainable Development</source>
  <publisher>Wiley Online Library</publisher>
  <pubdate>2015</pubdate>
  <volume>23</volume>
  <issue>3</issue>
  <fpage>176</fpage>
  <lpage>-187</lpage>
</bibl>

<bibl id="B13">
  <title><p>Sustainable Development Goals (SDGs): Are we successful in turning
  trade-offs into synergies?</p></title>
  <aug>
    <au><snm>Kroll</snm><fnm>C</fnm></au>
    <au><snm>Warchold</snm><fnm>A</fnm></au>
    <au><snm>Pradhan</snm><fnm>P</fnm></au>
  </aug>
  <source>Palgrave Communications</source>
  <publisher>Palgrave</publisher>
  <pubdate>2019</pubdate>
  <volume>5</volume>
  <issue>1</issue>
  <fpage>1</fpage>
  <lpage>-11</lpage>
</bibl>

<bibl id="B14">
  <title><p>Transforming our world: the 2030 Agenda for Sustainable
  Development</p></title>
  <aug>
    <au><cnm>{United Nations General Assembly}</cnm></au>
  </aug>
  <source>\url{https://sdgs.un.org/2030agenda}</source>
  <pubdate>2015</pubdate>
  <note>Accessed: 2021-02-23</note>
</bibl>

<bibl id="B15">
  <title><p>Gross polluters and vehicle emissions reduction</p></title>
  <aug>
    <au><snm>B\"{o}hm</snm><fnm>M</fnm></au>
    <au><snm>Nanni</snm><fnm>M</fnm></au>
    <au><snm>Pappalardo</snm><fnm>L</fnm></au>
  </aug>
  <source>Nature Sustainability</source>
  <pubdate>2022</pubdate>
</bibl>

<bibl id="B16">
  <title><p>A Deep Gravity model for mobility flows generation</p></title>
  <aug>
    <au><snm>Simini</snm><fnm>F</fnm></au>
    <au><snm>Barlacchi</snm><fnm>G</fnm></au>
    <au><snm>Luca</snm><fnm>M</fnm></au>
    <au><snm>Pappalardo</snm><fnm>L</fnm></au>
  </aug>
  <source>Nature Communications</source>
  <publisher>Nature Publishing Group</publisher>
  <pubdate>2021</pubdate>
  <volume>12</volume>
  <issue>1</issue>
  <fpage>1</fpage>
  <lpage>-13</lpage>
</bibl>

<bibl id="B17">
  <title><p>Gravity versus radiation models: On the importance of scale and
  heterogeneity in commuting flows</p></title>
  <aug>
    <au><snm>Masucci</snm><fnm>AP</fnm></au>
    <au><snm>Serras</snm><fnm>J</fnm></au>
    <au><snm>Johansson</snm><fnm>A</fnm></au>
    <au><snm>Batty</snm><fnm>M</fnm></au>
  </aug>
  <source>Physical Review E</source>
  <publisher>APS</publisher>
  <pubdate>2013</pubdate>
  <volume>88</volume>
  <issue>2</issue>
  <fpage>022812</fpage>
</bibl>

<bibl id="B18">
  <title><p>Principles of social science</p></title>
  <aug>
    <au><snm>Carey</snm><fnm>HC</fnm></au>
  </aug>
  <publisher>JB Lippincott \& Company</publisher>
  <pubdate>1867</pubdate>
  <volume>3</volume>
</bibl>

<bibl id="B19">
  <title><p>The P 1 P 2/D hypothesis: on the intercity movement of
  persons</p></title>
  <aug>
    <au><snm>Zipf</snm><fnm>GK</fnm></au>
  </aug>
  <source>American sociological review</source>
  <publisher>JSTOR</publisher>
  <pubdate>1946</pubdate>
  <volume>11</volume>
  <issue>6</issue>
  <fpage>677</fpage>
  <lpage>-686</lpage>
</bibl>

<bibl id="B20">
  <title><p>Systematic comparison of trip distribution laws and
  models</p></title>
  <aug>
    <au><snm>Lenormand</snm><fnm>M</fnm></au>
    <au><snm>Bassolas</snm><fnm>A</fnm></au>
    <au><snm>Ramasco</snm><fnm>JJ</fnm></au>
  </aug>
  <source>Journal of Transport Geography</source>
  <publisher>Elsevier</publisher>
  <pubdate>2016</pubdate>
  <volume>51</volume>
  <fpage>158</fpage>
  <lpage>-169</lpage>
</bibl>

<bibl id="B21">
  <title><p>The gravity model in transportation analysis: theory and
  extensions</p></title>
  <aug>
    <au><snm>Erlander</snm><fnm>S</fnm></au>
    <au><snm>Stewart</snm><fnm>NF</fnm></au>
  </aug>
  <publisher>Vsp</publisher>
  <pubdate>1990</pubdate>
  <volume>3</volume>
</bibl>

<bibl id="B22">
  <title><p>Modeling international mobility using roaming cell phone traces
  during COVID-19 pandemic</p></title>
  <aug>
    <au><snm>Luca</snm><fnm>M</fnm></au>
    <au><snm>Lepri</snm><fnm>B</fnm></au>
    <au><snm>Frias Martinez</snm><fnm>E</fnm></au>
    <au><snm>Lutu</snm><fnm>A</fnm></au>
  </aug>
  <source>EPJ Data Science</source>
  <publisher>Springer Berlin Heidelberg</publisher>
  <pubdate>2022</pubdate>
  <volume>11</volume>
  <issue>1</issue>
  <fpage>22</fpage>
</bibl>

<bibl id="B23">
  <title><p>A universal model for mobility and migration patterns</p></title>
  <aug>
    <au><snm>Simini</snm><fnm>F</fnm></au>
    <au><snm>Gonz{\'a}lez</snm><fnm>MC</fnm></au>
    <au><snm>Maritan</snm><fnm>A</fnm></au>
    <au><snm>Barab{\'a}si</snm><fnm>AL</fnm></au>
  </aug>
  <source>Nature</source>
  <pubdate>2012</pubdate>
  <volume>484</volume>
  <issue>7392</issue>
  <fpage>96</fpage>
  <lpage>-100</lpage>
</bibl>

<bibl id="B24">
  <title><p>{Gravity and scaling laws of city to city migration}</p></title>
  <aug>
    <au><snm>{Prieto Curiel}</snm><fnm>R</fnm></au>
    <au><snm>Pappalardo</snm><fnm>L</fnm></au>
    <au><snm>Gabrielli</snm><fnm>L</fnm></au>
    <au><snm>Bishop</snm><fnm>SR</fnm></au>
  </aug>
  <source>PLOS ONE</source>
  <publisher>Public Library of Science</publisher>
  <pubdate>2018</pubdate>
  <volume>13</volume>
  <issue>7</issue>
  <fpage>1</fpage>
  <lpage>-19</lpage>
  <url>https://doi.org/10.1371/journal.pone.0199892</url>
</bibl>

<bibl id="B25">
  <title><p>{Universal model of individual and population mobility on diverse
  spatial scales}</p></title>
  <aug>
    <au><snm>Yan</snm><fnm>XY</fnm></au>
    <au><snm>Wang</snm><fnm>WX</fnm></au>
    <au><snm>Gao</snm><fnm>ZY</fnm></au>
    <au><snm>Lai</snm><fnm>YC</fnm></au>
  </aug>
  <source>Nature Communications</source>
  <pubdate>2017</pubdate>
  <volume>8</volume>
  <issue>1</issue>
  <fpage>1639</fpage>
  <url>https://doi.org/10.1038/s41467-017-01892-8</url>
</bibl>

<bibl id="B26">
  <title><p>Generative Adversarial Nets</p></title>
  <aug>
    <au><snm>Goodfellow</snm><fnm>IJ</fnm></au>
    <au><snm>Pouget Abadie</snm><fnm>J</fnm></au>
    <au><snm>Mirza</snm><fnm>M</fnm></au>
    <au><snm>Xu</snm><fnm>B</fnm></au>
    <au><snm>Warde Farley</snm><fnm>D</fnm></au>
    <au><snm>Ozair</snm><fnm>S</fnm></au>
    <au><snm>Courville</snm><fnm>A</fnm></au>
    <au><snm>Bengio</snm><fnm>Y</fnm></au>
  </aug>
  <source>Proceedings of the 27th International Conference on Neural
  Information Processing Systems</source>
  <pubdate>2014</pubdate>
  <fpage>2672–2680</fpage>
</bibl>

<bibl id="B27">
  <title><p>Generative adversarial networks: An overview</p></title>
  <aug>
    <au><snm>Creswell</snm><fnm>A</fnm></au>
    <au><snm>White</snm><fnm>T</fnm></au>
    <au><snm>Dumoulin</snm><fnm>V</fnm></au>
    <au><snm>Arulkumaran</snm><fnm>K</fnm></au>
    <au><snm>Sengupta</snm><fnm>B</fnm></au>
    <au><snm>Bharath</snm><fnm>AA</fnm></au>
  </aug>
  <source>IEEE Signal Processing Magazine</source>
  <publisher>IEEE</publisher>
  <pubdate>2018</pubdate>
  <volume>35</volume>
  <issue>1</issue>
  <fpage>53</fpage>
  <lpage>-65</lpage>
</bibl>

<bibl id="B28">
  <title><p>Nips 2016 tutorial: Generative adversarial networks</p></title>
  <aug>
    <au><snm>Goodfellow</snm><fnm>I</fnm></au>
  </aug>
  <source>arXiv preprint arXiv:1701.00160</source>
  <pubdate>2016</pubdate>
</bibl>

<bibl id="B29">
  <title><p>Unsupervised Representation Learning with Deep Convolutional
  Generative Adversarial Networks</p></title>
  <aug>
    <au><snm>Radford</snm><fnm>A</fnm></au>
    <au><snm>Metz</snm><fnm>L</fnm></au>
    <au><snm>Chintala</snm><fnm>S</fnm></au>
  </aug>
  <pubdate>2016</pubdate>
</bibl>

<bibl id="B30">
  <title><p>trajGANs: Using generative adversarial networks for geo-privacy
  protection of trajectory data (Vision paper)</p></title>
  <aug>
    <au><snm>Liu</snm><fnm>X</fnm></au>
    <au><snm>Chen</snm><fnm>H</fnm></au>
    <au><snm>Andris</snm><fnm>C</fnm></au>
  </aug>
  <source>Location Privacy and Security Workshop</source>
  <pubdate>2018</pubdate>
  <fpage>1</fpage>
  <lpage>-7</lpage>
</bibl>

<bibl id="B31">
  <title><p>GANs based density distribution privacy-preservation on mobility
  data</p></title>
  <aug>
    <au><snm>Yin</snm><fnm>D</fnm></au>
    <au><snm>Yang</snm><fnm>Q</fnm></au>
  </aug>
  <source>Security and Communication Networks</source>
  <publisher>Hindawi</publisher>
  <pubdate>2018</pubdate>
  <volume>2018</volume>
</bibl>

<bibl id="B32">
  <title><p>Generative models for simulating mobility trajectories</p></title>
  <aug>
    <au><snm>Kulkarni</snm><fnm>V</fnm></au>
    <au><snm>Tagasovska</snm><fnm>N</fnm></au>
    <au><snm>Vatter</snm><fnm>T</fnm></au>
    <au><snm>Garbinato</snm><fnm>B</fnm></au>
  </aug>
  <source>arXiv preprint arXiv:1811.12801</source>
  <pubdate>2018</pubdate>
</bibl>

<bibl id="B33">
  <title><p>A Non-Parametric Generative Model for Human
  Trajectories.</p></title>
  <aug>
    <au><snm>Ouyang</snm><fnm>K</fnm></au>
    <au><snm>Shokri</snm><fnm>R</fnm></au>
    <au><snm>Rosenblum</snm><fnm>DS</fnm></au>
    <au><snm>Yang</snm><fnm>W</fnm></au>
  </aug>
  <source>IJCAI</source>
  <pubdate>2018</pubdate>
  <fpage>3812</fpage>
  <lpage>-3817</lpage>
</bibl>

<bibl id="B34">
  <title><p>A Variational Autoencoder Based Generative Model of Urban Human
  Mobility</p></title>
  <aug>
    <au><snm>{Huang}</snm><fnm>D.</fnm></au>
    <au><snm>{Song}</snm><fnm>X.</fnm></au>
    <au><snm>{Fan}</snm><fnm>Z.</fnm></au>
    <au><snm>{Jiang}</snm><fnm>R.</fnm></au>
    <au><snm>{Shibasaki}</snm><fnm>R.</fnm></au>
    <au><snm>{Zhang}</snm><fnm>Y.</fnm></au>
    <au><snm>{Wang}</snm><fnm>H.</fnm></au>
    <au><snm>{Kato}</snm><fnm>Y.</fnm></au>
  </aug>
  <source>2019 IEEE Conference on Multimedia Information Processing and
  Retrieval (MIPR)</source>
  <pubdate>2019</pubdate>
  <fpage>425</fpage>
  <lpage>430</lpage>
</bibl>

<bibl id="B35">
  <title><p>Learning to Simulate Human Mobility</p></title>
  <aug>
    <au><snm>Feng</snm><fnm>J</fnm></au>
    <au><snm>Yang</snm><fnm>Z</fnm></au>
    <au><snm>Xu</snm><fnm>F</fnm></au>
    <au><snm>Yu</snm><fnm>H</fnm></au>
    <au><snm>Wang</snm><fnm>M</fnm></au>
    <au><snm>Li</snm><fnm>Y</fnm></au>
  </aug>
  <source>Proceedings of the 26th ACM SIGKDD International Conference on
  Knowledge Discovery \& Data Mining</source>
  <publisher>New York, NY, USA: Association for Computing Machinery</publisher>
  <series><title><p>KDD '20</p></title></series>
  <pubdate>2020</pubdate>
  <fpage>3426–3433</fpage>
  <url>https://doi.org/10.1145/3394486.3412862</url>
</bibl>

<bibl id="B36">
  <title><p>Predicting taxi--passenger demand using streaming data</p></title>
  <aug>
    <au><snm>Moreira Matias</snm><fnm>L</fnm></au>
    <au><snm>Gama</snm><fnm>J</fnm></au>
    <au><snm>Ferreira</snm><fnm>M</fnm></au>
    <au><snm>Mendes Moreira</snm><fnm>J</fnm></au>
    <au><snm>Damas</snm><fnm>L</fnm></au>
  </aug>
  <source>IEEE Transactions on Intelligent Transportation Systems</source>
  <publisher>IEEE</publisher>
  <pubdate>2013</pubdate>
  <volume>14</volume>
  <issue>3</issue>
  <fpage>1393</fpage>
  <lpage>-1402</lpage>
</bibl>

<bibl id="B37">
  <title><p>A Mechanistic Data-Driven Approach to Synthesize Human Mobility
  Considering the Spatial, Temporal, and Social Dimensions Together</p></title>
  <aug>
    <au><snm>Cornacchia</snm><fnm>G</fnm></au>
    <au><snm>Pappalardo</snm><fnm>L</fnm></au>
  </aug>
  <source>ISPRS International Journal of Geo-Information</source>
  <pubdate>2021</pubdate>
  <volume>10</volume>
  <issue>9</issue>
</bibl>

<bibl id="B38">
  <title><p>Data-driven generation of spatio-temporal routines in human
  mobility</p></title>
  <aug>
    <au><snm>Pappalardo</snm><fnm>L</fnm></au>
    <au><snm>Simini</snm><fnm>F</fnm></au>
  </aug>
  <source>Data Mining and Knowledge Discovery</source>
  <pubdate>2018</pubdate>
  <volume>32</volume>
  <issue>3</issue>
  <fpage>787</fpage>
  <lpage>-829</lpage>
</bibl>

<bibl id="B39">
  <title><p>The TimeGeo modeling framework for urban mobility without travel
  surveys</p></title>
  <aug>
    <au><snm>Jiang</snm><fnm>S</fnm></au>
    <au><snm>Yang</snm><fnm>Y</fnm></au>
    <au><snm>Gupta</snm><fnm>S</fnm></au>
    <au><snm>Veneziano</snm><fnm>D</fnm></au>
    <au><snm>Athavale</snm><fnm>S</fnm></au>
    <au><snm>Gonzalez</snm><fnm>MC</fnm></au>
  </aug>
  <source>Proceedings of the National Academy of Sciences</source>
  <pubdate>2016</pubdate>
  <volume>113</volume>
  <fpage>201524261</fpage>
</bibl>

<bibl id="B40">
  <title><p>scikit-mobility: A Python Library for the Analysis, Generation, and
  Risk Assessment of Mobility Data</p></title>
  <aug>
    <au><snm>Pappalardo</snm><fnm>L</fnm></au>
    <au><snm>Simini</snm><fnm>F</fnm></au>
    <au><snm>Barlacchi</snm><fnm>G</fnm></au>
    <au><snm>Pellungrini</snm><fnm>R</fnm></au>
  </aug>
  <source>Journal of Statistical Software</source>
  <pubdate>2022</pubdate>
  <volume>103</volume>
  <issue>4</issue>
  <fpage>1</fpage>
  <lpage>-38</lpage>
</bibl>

<bibl id="B41">
  <title><p>CitiBike System Data</p></title>
  <aug>
    <au><cnm>citibikenyc.com</cnm></au>
  </aug>
  <pubdate>2013--</pubdate>
  <url>https://www.citibikenyc.com/system-data</url>
</bibl>

<bibl id="B42">
  <title><p>Divvy System Data</p></title>
  <aug>
    <au><cnm>divvybikes.com</cnm></au>
  </aug>
  <pubdate>2016--</pubdate>
  <url>https://www.divvybikes.com/system-data</url>
</bibl>

<bibl id="B43">
  <title><p>TLC Trip Record Data</p></title>
  <aug>
    <au><cnm>nyc.gov</cnm></au>
  </aug>
  <pubdate>2009--</pubdate>
  <url>https://www1.nyc.gov/site/tlc/about/tlc-trip-record-data.page</url>
</bibl>

<bibl id="B44">
  <title><p>TLC Trip Record Data</p></title>
  <aug>
    <au><cnm>chicago.gov</cnm></au>
  </aug>
  <pubdate>2013--</pubdate>
  <url>https://data.cityofchicago.org/Transportation/Taxi-Trips/wrvz-psew/data</url>
</bibl>

<bibl id="B45">
  <title><p>Comparing methods for comparing networks</p></title>
  <aug>
    <au><snm>Tantardini</snm><fnm>M</fnm></au>
    <au><snm>Ieva</snm><fnm>F</fnm></au>
    <au><snm>Tajoli</snm><fnm>L</fnm></au>
    <au><snm>Piccardi</snm><fnm>C</fnm></au>
  </aug>
  <source>Scientific reports</source>
  <publisher>Nature Publishing Group</publisher>
  <pubdate>2019</pubdate>
  <volume>9</volume>
  <issue>1</issue>
  <fpage>1</fpage>
  <lpage>-19</lpage>
</bibl>

<bibl id="B46">
  <title><p>{Systematic comparison of trip distribution laws and
  models}</p></title>
  <aug>
    <au><snm>Lenormand</snm><fnm>M</fnm></au>
    <au><snm>Bassolas</snm><fnm>A</fnm></au>
    <au><snm>Ramasco</snm><fnm>JJ</fnm></au>
  </aug>
  <source>Journal of Transport Geography</source>
  <pubdate>2016</pubdate>
  <volume>51</volume>
  <fpage>158</fpage>
  <lpage>-169</lpage>
  <url>https://www.sciencedirect.com/science/article/pii/S0966692315002422</url>
</bibl>

<bibl id="B47">
  <title><p>Cut based method for comparing complex networks</p></title>
  <aug>
    <au><snm>Liu</snm><fnm>Q</fnm></au>
    <au><snm>Dong</snm><fnm>Z</fnm></au>
    <au><snm>Wang</snm><fnm>E</fnm></au>
  </aug>
  <source>Scientific reports</source>
  <publisher>Nature Publishing Group</publisher>
  <pubdate>2018</pubdate>
  <volume>8</volume>
  <issue>1</issue>
  <fpage>1</fpage>
  <lpage>-11</lpage>
</bibl>

<bibl id="B48">
  <title><p>Approximating the cut-norm via Grothendieck's
  inequality</p></title>
  <aug>
    <au><snm>Alon</snm><fnm>N</fnm></au>
    <au><snm>Naor</snm><fnm>A</fnm></au>
  </aug>
  <source>Proceedings of the thirty-sixth annual ACM symposium on Theory of
  computing</source>
  <pubdate>2004</pubdate>
  <fpage>72</fpage>
  <lpage>-80</lpage>
</bibl>

<bibl id="B49">
  <title><p>An optimal SDP algorithm for Max-Cut, and equally optimal Long Code
  tests</p></title>
  <aug>
    <au><snm>O'Donnell</snm><fnm>R</fnm></au>
    <au><snm>Wu</snm><fnm>Y</fnm></au>
  </aug>
  <source>Proceedings of the fortieth annual ACM symposium on Theory of
  computing</source>
  <pubdate>2008</pubdate>
  <fpage>335</fpage>
  <lpage>-344</lpage>
</bibl>

<bibl id="B50">
  <title><p>cutnorm package</p></title>
  <aug>
    <au><snm>Chiu</snm><fnm>PK</fnm></au>
  </aug>
  <pubdate>2018--</pubdate>
  <url>https://pypi.org/project/cutnorm/</url>
</bibl>

<bibl id="B51">
  <title><p>Jensen-Shannon divergence and Hilbert space embedding</p></title>
  <aug>
    <au><snm>Fuglede</snm><fnm>B</fnm></au>
    <au><snm>Topsoe</snm><fnm>F</fnm></au>
  </aug>
  <source>International Symposium onInformation Theory, 2004. ISIT 2004.
  Proceedings.</source>
  <pubdate>2004</pubdate>
  <fpage>31</fpage>
</bibl>

<bibl id="B52">
  <title><p>Information theory and statistics</p></title>
  <aug>
    <au><snm>Kullback</snm><fnm>S</fnm></au>
  </aug>
  <publisher>Courier Corporation</publisher>
  <pubdate>1997</pubdate>
</bibl>

<bibl id="B53">
  <title><p>R{\'e}nyi divergence and Kullback-Leibler divergence</p></title>
  <aug>
    <au><snm>Van Erven</snm><fnm>T</fnm></au>
    <au><snm>Harremos</snm><fnm>P</fnm></au>
  </aug>
  <source>IEEE Transactions on Information Theory</source>
  <publisher>IEEE</publisher>
  <pubdate>2014</pubdate>
  <volume>60</volume>
  <issue>7</issue>
  <fpage>3797</fpage>
  <lpage>-3820</lpage>
</bibl>

<bibl id="B54">
  <title><p>Graph kernels</p></title>
  <aug>
    <au><snm>Vishwanathan</snm><fnm>S V N</fnm></au>
    <au><snm>Schraudolph</snm><fnm>NN</fnm></au>
    <au><snm>Kondor</snm><fnm>R</fnm></au>
    <au><snm>Borgwardt</snm><fnm>KM</fnm></au>
  </aug>
  <source>Journal of Machine Learning Research</source>
  <pubdate>2010</pubdate>
  <volume>11</volume>
  <fpage>1201</fpage>
  <lpage>-1242</lpage>
</bibl>

<bibl id="B55">
  <title><p>Graph kernels: A survey</p></title>
  <aug>
    <au><snm>Nikolentzos</snm><fnm>G</fnm></au>
    <au><snm>Siglidis</snm><fnm>G</fnm></au>
    <au><snm>Vazirgiannis</snm><fnm>M</fnm></au>
  </aug>
  <source>Journal of Artificial Intelligence Research</source>
  <pubdate>2021</pubdate>
  <volume>72</volume>
  <fpage>943</fpage>
  <lpage>-1027</lpage>
</bibl>

<bibl id="B56">
  <title><p>The graph neural network model</p></title>
  <aug>
    <au><snm>Scarselli</snm><fnm>F</fnm></au>
    <au><snm>Gori</snm><fnm>M</fnm></au>
    <au><snm>Tsoi</snm><fnm>AC</fnm></au>
    <au><snm>Hagenbuchner</snm><fnm>M</fnm></au>
    <au><snm>Monfardini</snm><fnm>G</fnm></au>
  </aug>
  <source>IEEE transactions on neural networks</source>
  <publisher>IEEE</publisher>
  <pubdate>2008</pubdate>
  <volume>20</volume>
  <issue>1</issue>
  <fpage>61</fpage>
  <lpage>-80</lpage>
</bibl>

<bibl id="B57">
  <title><p>Wasserstein generative adversarial networks</p></title>
  <aug>
    <au><snm>Arjovsky</snm><fnm>M</fnm></au>
    <au><snm>Chintala</snm><fnm>S</fnm></au>
    <au><snm>Bottou</snm><fnm>L</fnm></au>
  </aug>
  <source>International conference on machine learning</source>
  <pubdate>2017</pubdate>
  <fpage>214</fpage>
  <lpage>-223</lpage>
</bibl>

<bibl id="B58">
  <title><p>From gan to wgan</p></title>
  <aug>
    <au><snm>Weng</snm><fnm>L</fnm></au>
  </aug>
  <source>arXiv preprint arXiv:1904.08994</source>
  <pubdate>2019</pubdate>
</bibl>

<bibl id="B59">
  <title><p>Improved Training of Wasserstein GANs</p></title>
  <aug>
    <au><snm>Gulrajani</snm><fnm>I</fnm></au>
    <au><snm>Ahmed</snm><fnm>F</fnm></au>
    <au><snm>Arjovsky</snm><fnm>M</fnm></au>
    <au><snm>Dumoulin</snm><fnm>V</fnm></au>
    <au><snm>Courville</snm><fnm>A</fnm></au>
  </aug>
  <source>Proceedings of the 31st International Conference on Neural
  Information Processing Systems</source>
  <pubdate>2017</pubdate>
  <fpage>5769–5779</fpage>
</bibl>

<bibl id="B60">
  <title><p>Generating synthetic mobility data for a realistic population with
  RNNs to improve utility and privacy</p></title>
  <aug>
    <au><snm>Berke</snm><fnm>A</fnm></au>
    <au><snm>Doorley</snm><fnm>R</fnm></au>
    <au><snm>Larson</snm><fnm>K</fnm></au>
    <au><snm>Moro</snm><fnm>E</fnm></au>
  </aug>
  <source>CoRR</source>
  <pubdate>2022</pubdate>
  <volume>abs/2201.01139</volume>
  <url>https://arxiv.org/abs/2201.01139</url>
</bibl>

<bibl id="B61">
  <title><p>The purpose of motion: Learning activities from Individual Mobility
  Networks</p></title>
  <aug>
    <au><snm>{Rinzivillo}</snm><fnm>S.</fnm></au>
    <au><snm>{Gabrielli}</snm><fnm>L.</fnm></au>
    <au><snm>{Nanni}</snm><fnm>M.</fnm></au>
    <au><snm>{Pappalardo}</snm><fnm>L.</fnm></au>
    <au><snm>{Pedreschi}</snm><fnm>D.</fnm></au>
    <au><snm>{Giannotti}</snm><fnm>F.</fnm></au>
  </aug>
  <source>2014 International Conference on Data Science and Advanced Analytics
  (DSAA)</source>
  <pubdate>2014</pubdate>
  <fpage>312</fpage>
  <lpage>318</lpage>
</bibl>

<bibl id="B62">
  <title><p>Unravelling daily human mobility motifs</p></title>
  <aug>
    <au><snm>Schneider</snm><fnm>CM</fnm></au>
    <au><snm>Belik</snm><fnm>V</fnm></au>
    <au><snm>Couronné</snm><fnm>T</fnm></au>
    <au><snm>Smoreda</snm><fnm>Z</fnm></au>
    <au><snm>González</snm><fnm>MC</fnm></au>
  </aug>
  <source>Journal of The Royal Society Interface</source>
  <pubdate>2013</pubdate>
  <volume>10</volume>
  <issue>84</issue>
  <fpage>20130246</fpage>
</bibl>

<bibl id="B63">
  <title><p>Deep learning</p></title>
  <aug>
    <au><snm>Goodfellow</snm><fnm>I</fnm></au>
    <au><snm>Bengio</snm><fnm>Y</fnm></au>
    <au><snm>Courville</snm><fnm>A</fnm></au>
  </aug>
  <publisher>MIT press</publisher>
  <pubdate>2016</pubdate>
</bibl>

<bibl id="B64">
  <title><p>Generative adversarial networks in computer vision: A survey and
  taxonomy</p></title>
  <aug>
    <au><snm>Wang</snm><fnm>Z</fnm></au>
    <au><snm>She</snm><fnm>Q</fnm></au>
    <au><snm>Ward</snm><fnm>TE</fnm></au>
  </aug>
  <source>ACM Computing Surveys (CSUR)</source>
  <publisher>ACM New York, NY, USA</publisher>
  <pubdate>2021</pubdate>
  <volume>54</volume>
  <issue>2</issue>
  <fpage>1</fpage>
  <lpage>-38</lpage>
</bibl>

<bibl id="B65">
  <title><p>Conditional generative adversarial nets</p></title>
  <aug>
    <au><snm>Mirza</snm><fnm>M</fnm></au>
    <au><snm>Osindero</snm><fnm>S</fnm></au>
  </aug>
  <source>arXiv preprint arXiv:1411.1784</source>
  <pubdate>2014</pubdate>
</bibl>

</refgrp>
} 






\end{backmatter}
\end{document}